%% file: main.tex
\documentclass{article}

\usepackage[accepted]{icml2020}
\input{preamble}

\icmltitlerunning{SoftSort: A Continuous Relaxation for the \texttt{argsort} Operator}

\begin{document}

\twocolumn[
\icmltitle{SoftSort: A Continuous Relaxation for the \texttt{argsort} Operator}



\icmlsetsymbol{equal}{*}

\begin{icmlauthorlist}
\icmlauthor{Sebastian Prillo}{equal,ucb}
\icmlauthor{Julian Martin Eisenschlos}{equal,goo}
\end{icmlauthorlist}

\icmlaffiliation{ucb}{University of California, Berkeley, California, USA}
\icmlaffiliation{goo}{Google Research, Zurich, Switzerland}

\icmlcorrespondingauthor{Sebastian Prillo}{sprillo@berkeley.edu}
\icmlcorrespondingauthor{Julian Martin Eisenschlos}{eisenjulian@google.com}

\icmlkeywords{Machine Learning, sorting, ranking, ICML}

\vskip 0.3in
]



\begin{NoHyper}
\printAffiliationsAndNotice{\icmlEqualContribution} 
\end{NoHyper}

\begin{abstract}
While sorting is an important procedure in computer science, the \texttt{argsort} operator - which takes as input a vector and returns its sorting permutation - has a discrete image and thus zero gradients almost everywhere.
This prohibits end-to-end, gradient-based learning of models that rely on the \texttt{argsort} operator.
A natural way to overcome this problem is to replace the \texttt{argsort} operator with a continuous relaxation.
Recent work has shown a number of ways to do this, but the relaxations proposed so far are computationally complex.
In this work we propose a simple continuous relaxation for the \texttt{argsort} operator which has the following qualities: it can be implemented in three lines of code, achieves state-of-the-art performance, is easy to reason about mathematically - substantially simplifying proofs - and is faster than competing approaches.
We open source the code to reproduce all of the experiments and results.
\end{abstract}

\section{Introduction}

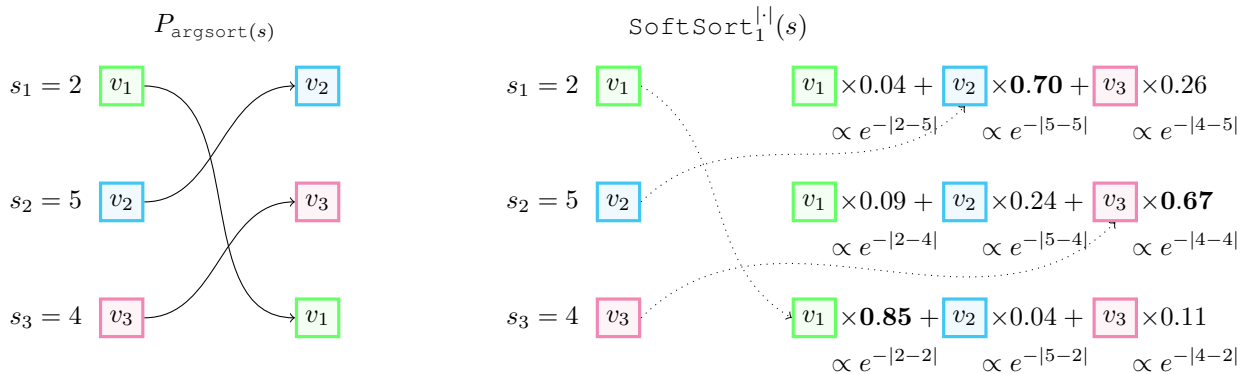
\begin{figure*}[ht]
\begin{center}

\begin{tikzpicture}[
green_node/.style={rectangle, draw=green!60, fill=green!5, very thick, minimum size=5mm},
cyan_node/.style={rectangle, draw=cyan!60, fill=cyan!5, very thick, minimum size=5mm},
magenta_node/.style={rectangle, draw=magenta!60, fill=magenta!5, very thick, minimum size=5mm},
big_green_node/.style={rectangle, draw=green!60, fill=green!5, very thick, minimum size=5mm},
big_cyan_node/.style={rectangle, draw=cyan!60, fill=cyan!5, very thick, minimum size=5mm},
big_magenta_node/.style={rectangle, draw=magenta!60, fill=magenta!5, very thick, minimum size=7mm},
link/.style={white, double=black, line width = 1.8pt, double distance=0.8pt, ->},
]
\node[green_node]   (b1)            {$v_1$};
\node[cyan_node]    (b2)            [below=of b1] {$v_2$};
\node[magenta_node] (b3)            [below=of b2] {$v_3$};

\node [above right=0.2cm and -0.0cm of b1] {$P_{\texttt{argsort}(s)}$};

\node [left=0.1cm of b1] {$s_1=2$};
\node [left=0.1cm of b2] {$s_2=5$};
\node [left=0.1cm of b3] {$s_3=4$};

\node[cyan_node]    (bb2)           [right=2cm of b1] {$v_2$};
\node[magenta_node] (bb3)           [right=2cm of b2] {$v_3$};
\node[green_node]   (bb1)           [right=2cm of b3] {$v_1$};

\draw[->] (b1.east) to[out=0,in=180] (bb1.west);
\draw[->] (b2.east) to[out=0,in=180] (bb2.west);
\draw[->] (b3.east) to[out=0,in=180] (bb3.west);

\node[green_node]   (v1)            [right=6cm of b1]{$v_1$};
\node[cyan_node]    (v2)            [below=of v1] {$v_2$};
\node[magenta_node] (v3)            [below=of v2] {$v_3$};

\node [above right=0.2cm and -0.3cm of v1] {$\SimpleSort{}^{|\cdot|}_{1}(s)$};

\node [left=0.1cm of v1] {$s_1=2$};
\node [left=0.1cm of v2] {$s_2=5$};
\node [left=0.1cm of v3] {$s_3=4$};

\node[green_node]   (vv11)           [right=2cm of v1] {$v_1$}; \node[right=-0.1cm of vv11] {$\times 0.04 \ +$};
\node[below right=-0.0cm and -0.2cm of vv11] {$\propto e^{-|2-5|}$};
\node[cyan_node]    (vv12)           [right=4cm of v1] {$v_2$}; \node[right=-0.1cm of vv12] {$\times \mathbf{0.70} \ +$};
\node[below right=-0.0cm and -0.2cm of vv12] {$\propto e^{-|5-5|}$};
\node[magenta_node] (vv13)           [right=6cm of v1] {$v_3$}; \node[right=-0.1cm of vv13] {$\times 0.26$};
\node[below right=-0.0cm and -0.2cm of vv13] {$\propto e^{-|4-5|}$};

\node[green_node]   (vv21)           [right=2cm of v2] {$v_1$}; \node[right=-0.1cm of vv21] {$\times 0.09 \ +$};
\node[below right=-0.0cm and -0.2cm of vv21] {$\propto e^{-|2-4|}$};
\node[cyan_node]    (vv22)           [right=4cm of v2] {$v_2$}; \node[right=-0.1cm of vv22] {$\times 0.24 \ +$};
\node[below right=-0.0cm and -0.2cm of vv22] {$\propto e^{-|5-4|}$};
\node[magenta_node] (vv23)           [right=6cm of v2] {$v_3$}; \node[right=-0.1cm of vv23] {$\times \mathbf{0.67}$};
\node[below right=-0.0cm and -0.2cm of vv23] {$\propto e^{-|4-4|}$};

\node[big_green_node]   (vv31)           [right=2cm of v3] {$v_1$}; \node[right=-0.1cm of vv31] {$\times \mathbf{0.85} \ +$};
\node[below right=-0.0cm and -0.2cm of vv31] {$\propto e^{-|2-2|}$};
\node[cyan_node]    (vv32)           [right=4cm of v3] {$v_2$}; \node[right=-0.1cm of vv32] {$\times 0.04 \ +$};
\node[below right=-0.0cm and -0.2cm of vv32] {$\propto e^{-|5-2|}$};
\node[magenta_node] (vv33)           [right=6cm of v3] {$v_3$}; \node[right=-0.1cm of vv33] {$\times 0.11$};
\node[below right=-0.0cm and -0.2cm of vv33] {$\propto e^{-|4-2|}$};

\draw[->,dotted] (v1.east) to[out=-30,in=150] (vv31.west);
\draw[->,dotted] (v2.east) to[out=45,in=-135] (vv12.south);
\draw[->,dotted] (v3.east) to[out=45,in=-135] (vv23.south);
\end{tikzpicture}

\caption{Left: Standard $P_\texttt{argsort}$ operation of real value scores $s_i$ on the corresponding vectors $v_i$. The output is a permutation of the vectors to match the decreasing order of the scores $s_i$. Right: \SimpleSort{} operator applied to the same set of scores and vectors. The output is now a sequence of convex combinations of the vectors that approximates the one on the left and is differentiable with respect to $s$.}
\label{fig:simplesort_plots}
\end{center}
\vskip -0.2in
\end{figure*}

Gradient-based optimization lies at the core of the success of deep learning. However, many common operators have discrete images and thus exhibit zero gradients almost everywhere, making them unsuitable for gradient-based optimization. Examples include the \emph{Heaviside step} function, the \texttt{argmax} operator, and - the central object of this work - the \texttt{argsort} operator. To overcome this limitation, continuous relaxations for these operators can be constructed. For example, the sigmoid function serves as a continuous relaxation for the \emph{Heaviside step} function, and the \texttt{softmax} operator serves as a continuous relaxation for the \texttt{argmax}. These continuous relaxations have the crucial property that they provide meaningful gradients that can drive optimization. Because of this, operators such as the \texttt{softmax} are ubiquitous in deep learning. In this work we are concerned with continuous relaxations for the \texttt{argsort} operator.

Formally, we define the \texttt{argsort} operator as the mapping $\texttt{argsort} : \mathbb R^n \rightarrow \mathcal S_n$ from $n$-dimensional real vectors $s \in \mathbb R^n$ to the set of permutations over $n$ elements $\mathcal S_n \subseteq \{1,2,\dots,n\}^n$, where $\texttt{argsort}(s)$ returns the permutation that sorts $s$ in decreasing order\footnote{This is called the \texttt{sort} operator in \cite{2018_grover}. We adopt the more conventional naming.}. For example, for the input vector $s = [9,1,5,2]^T$ we have $\texttt{argsort}(s) = [1, 3, 4, 2]^T$. If we let $\mathcal P_n \subseteq \{0, 1\}^{n \times n} \subset \mathbb R^{n \times n}$ denote the set of permutation matrices of dimension $n$, following \cite{2018_grover} we can define, for a permutation $\pi \in \mathcal S_n$, the permutation matrix $P_\pi \in \mathcal P_n$ as:
\begin{equation*}
P_\pi[i, j] =
\begin{cases}
1\text{ if }j = \pi_i,\\
0\text{ otherwise}
\end{cases}
\end{equation*}
This is simply the one-hot representation of $\pi$.
Note that with these definitions, the mapping $\texttt{sort} : \mathbb R^n \rightarrow \mathbb R^n$ that sorts $s$ in decreasing order is $\texttt{sort}(s) = P_{\texttt{argsort}(s)}s$.
Also, if we let $\bar{1}_n = [1,2,\dots,n]^T$, then the \texttt{argsort} operator can be recovered from $P_{\texttt{argsort}(\cdot)} : \mathbb R^n \rightarrow \mathcal P_n$ by a simple matrix multiplication via
\[
\texttt{argsort}(s) = P_{\texttt{argsort}(s)}\bar{1}_n
\]
Because of this reduction from the \texttt{argsort} operator to the $P_{\texttt{argsort}(\cdot)}$ operator, in this work, as in previous works \cite{2018_mena, 2018_grover, 2019_cuturi}, our strategy to derive a continuous relaxation for the \texttt{argsort} operator is to instead derive a continuous relaxation for the $P_{\texttt{argsort}(\cdot)}$ operator. This is analogous to the way that the \texttt{softmax} operator relaxes the \texttt{argmax} operator.

The main contribution of this paper is the proposal of a family of simple continuous relaxation for the $P_{\texttt{argsort}(\cdot)}$ operator, which we call $\SimpleSort{}$, and define as follows:
\begin{equation*}\label{eq:simplesort}
\SimpleSort{}^d_{\tau}(s) = \texttt{softmax}\left(\frac{-d\left(\texttt{sort}(s) \mathds 1^T, \mathds 1 s^T\right)}{\tau}\right)
\end{equation*}
where the softmax operator is applied row-wise, $d$ is any differentiable almost everywhere, semi--metric function of $\mathbb R$ applied pointwise, and $\tau$ is a temperature parameter that controls the degree of the approximation. In simple words: \textit{the $r$-th row of the \SimpleSort{} operator is the \texttt{softmax} of the negative distances to the $r$-th largest element}.

Throughout this work we will predominantly use $d(x, y) = |x - y|$ (the $L_1$ norm), but our theoretical results hold for any such $d$, making the approach flexible. The \SimpleSort{} operator is trivial to implement in automatic differentiation libraries such as TensorFlow \cite{2016_abadi} and PyTorch \cite{2017_paszke}, and we show that:
\begin{itemize}
\item \SimpleSort{} achieves state-of-the-art performance on multiple benchmarks that involve reordering elements.
\item \SimpleSort{} shares the same desirable properties as the \texttt{NeuralSort} operator \cite{2018_grover}. Namely, it is row-stochastic, converges to $P_{\texttt{argsort}(\cdot)}$ in the limit of the temperature, and can be projected onto a permutation matrix using the row-wise \texttt{argmax} operation. However, \SimpleSort{} is significantly easier to reason about mathematically, which substantially simplifies the proof of these properties.
\item The \SimpleSort{} operator is faster than the \texttt{NeuralSort} operator of \cite{2018_grover}, the fastest competing approach, and empirically just as easy to optimize in terms of the number of gradient steps required for the training objective to converge.
\end{itemize}

Therefore, the \SimpleSort{} operator advances the state of the art in differentiable sorting by significantly simplifying previous approaches. To better illustrate the usefulness of the mapping defined by \SimpleSort{}, we show in Figure~\ref{fig:simplesort_plots} the result of applying the operator to soft-sort a sequence of vectors $v_i$ $(1\leq i \leq n)$ according to the order given by respective scores $s_i \in \mathbb R$. Soft-sorting the $v_i$ is achieved by multiplying to the left by $\SimpleSort(s)$.

The code and experiments are available at \url{https://github.com/sprillo/softsort}

\section{Related Work}

Relaxed operators for sorting procedures were first proposed in the context of Learning to Rank with the end goal of enabling direct optimization of Information Retrieval (IR) metrics. Many IR metrics, such as the Normalized Discounted Cumulative Gain (NDCG) \cite{2002_kalervo}, are defined in terms of \textit{ranks}. Formally, the \texttt{rank} operator is defined as the function $\texttt{rank} : \mathbb R^n \rightarrow \mathcal S_n$ that returns the inverse of the \texttt{argsort} permutation: $\texttt{rank}(s) = \texttt{argsort}(s)^{-1}$, or equivalently $P_{\texttt{rank}(s)} = P_{\texttt{argsort}(s)}^T$. The $\texttt{rank}$ operator is thus intimately related to the $\texttt{argsort}$ operator; in fact, a relaxation for the $P_{\texttt{rank}(\cdot)}$ operator can be obtained by transposing a relaxation for the $P_{\texttt{argsort}(\cdot)}$ operator, and vice-versa; this duality is apparent in the work of \cite{2019_cuturi}.

We begin by discussing previous work on relaxed \texttt{rank} operators in section~\ref{sec:relaxed_rank_operators}. Next, we discuss more recent work, which deals with relaxations for the $P_{\texttt{argsort}(\cdot)}$ operator.

\subsection{Relaxed Rank Operators}\label{sec:relaxed_rank_operators}

The first work to propose a relaxed \texttt{rank} operator is that of \cite{2008_taylor}. The authors introduce the relaxation $\texttt{SoftRank}_\tau : \mathbb R^n \rightarrow \mathbb R^n$ given by $\texttt{SoftRank}_\tau(s) = \mathbb E[\texttt{rank}(s + z)]$ where $z \sim \mathcal N_n(0, \tau I_n)$, and show that this relaxation, as well as its gradients, can be computed exactly in time $\mathcal O(n^3)$. Note that as $\tau \rightarrow 0$, $\texttt{SoftRank}_\tau(s) \rightarrow \texttt{rank}(s)$\footnote{Except when there are ties, which we assume is not the case. Ties are merely a technical nuisance and do not represent a problem for any of the methods (ours or other's) discussed in this paper.}. This relaxation is used in turn to define a surrogate for NDCG which can be optimized directly.

In \cite{2010_qin}, another relaxation for the \texttt{rank} operator $\widehat{\pi}_\tau : \mathbb R^n \rightarrow \mathbb R^n$ is proposed, defined as:
\begin{equation}\label{eqn:2010_qin}
\widehat{\pi}_\tau(s)[i] = 1 + \sum_{j \neq i} \sigma\left(\frac{s_i - s_j}{\tau}\right)
\end{equation}
where $\sigma(x) = (1 + \exp\{-x\})^{-1}$ is the sigmoid function. Again, $\widehat{\pi}_\tau(s) \rightarrow \texttt{rank}(s)$ as $\tau \rightarrow 0$. This operator can be computed in time $\mathcal{O}(n^2)$, which is faster than the $\mathcal O(n^3)$ approach of \cite{2008_taylor}. 

Note that the above two approaches cannot be used to define a relaxation for the \texttt{argsort} operator. Indeed, $\texttt{SoftRank}_\tau(s)$ and $\widehat{\pi}_\tau(s)$ are not relaxations for the $P_{\texttt{rank}(\cdot)}$ operator. Instead, they directly relax the $\texttt{rank}$ operator, and there is no easy way to obtain a relaxed \texttt{argsort} or $P_{\texttt{argsort}(\cdot)}$ operator from them.

\subsection{Sorting via Bipartite Maximum Matchings}

The work of \cite{2018_mena} draws an analogy between the \texttt{argmax} operator and the \texttt{argsort} operator by means of bipartite maximum matchings: the \texttt{argmax} operator applied to an $n$-dimensional vector $s$ can be viewed as a maximum matching on a bipartite graph with $n$ vertices in one component and $1$ vertex in the other component, the edge weights equal to the given vector $s$; similarly, a permutation matrix can be seen as a maximum matching on a bipartite graph between two groups of $n$ vertices with edge weights given by a matrix $X \in \mathbb R^{n \times n}$. This induces a mapping $M$ (for `matching') from the set of square matrices $X \in \mathbb R^{n \times n}$ to the set $\mathcal P_n$. Note that this mapping has arity $M : \mathbb R^{n \times n} \rightarrow \mathcal P_n$, unlike the $P_{\texttt{argsort}(\cdot)}$ operator which has arity $P_{\texttt{argsort}(\cdot)} : \mathbb R^n \rightarrow \mathcal P_n$.


Like the $P_{\texttt{argsort}(\cdot)}$ operator, the $M$ operator has discrete image $\mathcal P_n$, so to enable end-to-end gradient-based optimization, \cite{2018_mena} propose to relax the matching operator $M(X)$ by means of the Sinkhorn operator $S(X / \tau)$; $\tau$ is a temperature parameter that controls the degree of the approximation; as $\tau \rightarrow 0$ they show that $S(X / \tau) \rightarrow M(X)$. The Sinkhorn operator $S$ maps the square matrix $X / \tau$ to the Birkhoff polytope $\mathcal B_n$, which is defined as the set of doubly stochastic matrices (i.e. rows and columns summing to $1$).

The computational complexity of the \cite{2018_mena} approach to differentiable sorting is thus $\mathcal O(Ln^2)$ where $L$ is the number of Sinkhorn iterates used to approximate $S(X / \tau)$; the authors use $L = 20$ for all experiments.

\subsection{Sorting via Optimal Transport}

The recent work of \cite{2019_cuturi} also makes use of the Sinkhorn operator to derive a continuous relaxation for the $P_{\texttt{argsort}(\cdot)}$ operator. This time, the authors are motivated by the observation that a sorting permutation for $s \in \mathbb R^n$ can be recovered from an optimal transport plan between two discrete measures defined on the real line, one of which is supported on the elements of $s$ and the other of which is supported on arbitrary values $y_1 < \dots < y_n$. Indeed, the optimal transport plan between the probability measures $\frac{1}{n} \sum_{i = 1}^n \delta_{s_i}$ and $\frac{1}{n} \sum_{i = 1}^n \delta_{y_i}$ (where $\delta_x$ is the Dirac delta at $x$) is given by the matrix $P_{\texttt{argsort}(s)}^T$.

Notably, a variant of the optimal transport problem with entropy regularization yields instead a continuous relaxation $P^\epsilon_{\texttt{argsort}(s)}$ mapping $s$ to the Birkhoff polytope $\mathcal B_n$; $\epsilon$ plays a role similar to the temperature in \cite{2018_mena}, with $P^\epsilon_{\texttt{argsort}(s)} \rightarrow P_{\texttt{argsort}(s)}$ as $\epsilon \rightarrow 0$. This relaxation can be computed via Sinkhorn iterates, and enables the authors to relax $P_{\texttt{argsort}(\cdot)}$ by means of $P^\epsilon_{\texttt{argsort}(s)}$. Gradients can be computed by backpropagating through the Sinkhorn operator as in \cite{2018_mena}.

The computational complexity of this approach is again $\mathcal O(Ln^2)$. However, the authors show that a generalization of their method can be used to compute relaxed quantiles in time $\mathcal O(Ln)$, which is interesting in its own right.

\subsection{Sorting via a sum-of-top-k elements identity}

Finally, a more computationally efficient approach to differentiable sorting is proposed in \cite{2018_grover}. The authors rely on an identity that expresses the sum of the top $k$ elements of a vector $s \in \mathbb R^n$ as a symmetric function of $s_1,\dots,s_n$ that only involves \texttt{max} and \texttt{min} operations~\citep[Lemma 1]{2003_ogryczak}. Based on this identity, and denoting by $A_s$ the matrix of \textit{absolute} pairwise differences of elements of $s$, namely $A_s[i, j] = |s_i - s_j|$, the authors prove the identity:
\begin{equation} \label{eq:2018_grover_eq4}
P_{\texttt{argsort}(s)}[i, j] =
\begin{cases}
1\text{ if }j = \texttt{argmax}(c_i),\\
0\text{ otherwise}
\end{cases}
\end{equation}
where $c_i = (n + 1 - 2i)s - A_s \mathds 1$, and $\mathds 1$ denotes the column vector of all ones.

Therefore, by replacing the $\texttt{argmax}$ operator in Eq.~\ref{eq:2018_grover_eq4} by a row-wise \texttt{softmax}, the authors arrive at the following continuous relaxation for the $P_{\texttt{argsort}(\cdot)}$ operator, which they call \texttt{NeuralSort}:
\begin{equation} \label{eq:2018_grover_eq5}
\texttt{NeuralSort}_\tau(s)[i, :] = \texttt{softmax}\left(\frac{c_i}{\tau}\right)
\end{equation}
Again, $\tau > 0$ is a temperature parameter that controls the degree of the approximation; as $\tau \rightarrow 0$ they show that $\texttt{NeuralSort}_\tau(s) \rightarrow P_{\texttt{argsort}(s)}$. Notably, the relaxation proposed by \cite{2018_grover} can be computed in time $\mathcal O(n^2)$, making it much faster than the competing approaches of \cite{2018_mena, 2019_cuturi}.

\section{\SimpleSort{}: A simple relaxed sorting operator}

\begin{figure*}[ht]
\begin{center}
\begin{equation*}
\texttt{NeuralSort}_\tau(s) = g_\tau
\begin{pmatrix}
0                        & s_2 - s_1         & 3s_3 - s_1 - 2s_2 & 5s_4 - s_1 - 2s_2 - 2s_3 \\
s_2 - s_1                & 0                 & s_3 - s_2         & 3s_4 - s_2 - 2s_3 \\
2s_2 + s_3 - 3s_1        & s_3 - s_2         & 0                 & s_4 - s_3 \\
2s_2 + 2s_3 + s_4 - 5s_1 & 2s_3 + s_4 - 3s_2 & s_4 - s_3         & 0 \\
\end{pmatrix}
\end{equation*}
\caption{The $4$-dimensional \texttt{NeuralSort} operator on the region of space where $s_1 \ge s_2 \ge s_3 \ge s_4$. We define $g_\tau(X)$ as a row-wise \texttt{softmax} with temperature: $\texttt{softmax}(X / \tau)$. As the most closely related work, this is the main baseline for our experiments.}
\label{eq:neuralsort_logits}
\end{center}
\vskip -0.2in
\end{figure*}

\begin{figure*}[ht]
\begin{center}
\begin{equation*}
\SimpleSort{}^{|\cdot|}_{\tau}(s) = g_\tau
\begin{pmatrix}
0         & s_2 - s_1 & s_3 - s_1 & s_4 - s_1 \\
s_2 - s_1 & 0         & s_3 - s_2 & s_4 - s_2 \\
s_3 - s_1 & s_3 - s_2 & 0         & s_4 - s_3 \\
s_4 - s_1 & s_4 - s_2 & s_4 - s_3 & 0         \\
\end{pmatrix}
\end{equation*}
\caption{The $4$-dimensional \SimpleSort{} operator on the region of space where $s_1 \ge s_2 \ge s_3 \ge s_4$ using $d=|\cdot|$. We define $g_\tau(X)$ as a row-wise \texttt{softmax} with temperature: $\texttt{softmax}(X / \tau)$. Our formulation is much simpler than previous approaches.}
\label{eq:simplesort_logits}
\end{center}
\vskip -0.2in
\end{figure*}

In this paper we propose \SimpleSort{}, a simple continuous relaxation for the $P_{\texttt{argsort}(\cdot)}$ operator. We define \SimpleSort{} as follows:
\begin{equation}\label{eq:simplesort2}
\SimpleSort{}^d_{\tau}(s) = \texttt{softmax}\left(\frac{-d\left(\texttt{sort}(s) \mathds 1^T, \mathds 1 s^T\right)}{\tau}\right)
\end{equation}
where $\tau > 0$ is a temperature parameter that controls the degree of the approximation and $d$ is semi--metric function applied pointwise that is differentiable almost everywhere.
Recall that a semi--metric has all the properties of a metric except the triangle inequality, which is not required. Examples of semi--metrics in $\mathbb R$ include any positive power of the absolute value. The \SimpleSort{} operator has similar desirable properties to those of the \texttt{NeuralSort} operator, while being significantly simpler. Here we state and prove these properties. We start with the definition of \textit{Unimodal Row Stochastic Matrices} \cite{2018_grover}, which summarizes the properties of our relaxed operator:

\begin{definition} (Unimodal Row Stochastic Matrices).
An $n \times n$ matrix is Unimodal Row Stochastic (URS) if it satisfies the following conditions:
\begin{enumerate}
\item \textbf{Non-negativity:} $U[i, j] \ge 0\quad \forall i,j \in \{1,2,\dots,n\}$.
\item \textbf{Row Affinity:} $\sum_{j = 1}^n U[i, j] = 1\quad \forall i \in \{1,2,\dots,n\}$.
\item \textbf{Argmax Permutation:} Let $u$ denote a vector of size $n$ such that $u_i = \arg\max_j U[i, j]\quad \forall i \in \{1,2,\dots,n\}$. Then, $u \in \mathcal S_n$, i.e., it is a valid permutation.
\end{enumerate}
\end{definition}

While \texttt{NeuralSort} and \texttt{SoftSort} yield URS matrices (we will prove this shortly), the approaches of \cite{2018_mena, 2019_cuturi} yield bistochastic matrices. It is natural to ask whether URS matrices should be preferred over bistochastic matrices for relaxing the $P_{\texttt{argsort}(\cdot)}$ operator. Note that URS matrices are not comparable to bistochastic matrices: they drop the column-stochasticity condition, but require that each row have a distinct \texttt{argmax}, which is not true of bistochastic matrices. This means that URS matrices can be trivially projected onto the probability simplex, which is useful for e.g. straight-through gradient optimization, or whenever hard permutation matrices are required, such as at test time. The one property URS matrices lack is column-stochasticity, but this is not central to soft sorting. Instead, this property arises in the work of \cite{2018_mena} because their goal is to relax the bipartite matching operator (rather than the \texttt{argsort} operator), and in this context bistochastic matrices are the natural choice. Similarly, \cite{2019_cuturi} yields bistochastic matrices because they are the solutions to optimal transport problems (this does, however, allow them to simultaneously relax the \texttt{argsort} and \texttt{rank} operators). Since our only goal (as in the \texttt{NeuralSort} paper) is to relax the \texttt{argsort} operator, column-stochasticity can be dropped, and URS matrices are the more natural choice.


Now on to the main Theorem, which shows that \SimpleSort{} has the same desirable properties as \texttt{NeuralSort}. These are \citep[Theorem~4]{2018_grover}:

\begin{theorem}\label{main_theorem}
The \SimpleSort{} operator has the following properties:
\begin{enumerate}
\item Unimodality: $\forall \tau > 0$, $\SimpleSort{}^d_\tau(s)$ is a unimodal row stochastic matrix. Further, let $u$ denote the permutation obtained by applying $\texttt{argmax}$ row-wise to $\SimpleSort{}^d_\tau(s)$. Then, $u = \texttt{argsort}(s)$.
\item Limiting behavior: If no elements of $s$ coincide, then:
\begin{equation*}
    \lim_{\tau \rightarrow 0^+} \SimpleSort{}^d_\tau(s) = P_{\texttt{argsort}(s)}
\end{equation*}
In particular, this limit holds almost surely if the entries of $s$ are drawn from a distribution that is absolutely continuous w.r.t. the Lebesgue measure on $\mathbb R$.
\end{enumerate}
\end{theorem}


\textbf{Proof}.
\begin{enumerate}
\item Non-negativity and row affinity follow from the fact that every row in $\SimpleSort{}^d_\tau(s)$ is the result of a \texttt{softmax} operation. For the third property, we use that \texttt{softmax} preserves maximums and that $d(\cdot, x)$ has a unique minimum at $x$ for every $x\in\mathbb R$. Formally, let $\texttt{sort}(s) = [s_\ord{1}, \dots, s_\ord{n}]^T$, i.e. $s_\ord{1} \ge \dots \ge s_\ord{n}$ are the decreasing order statistics of $s$. Then:
\begin{align*}
u_i &= \argmax_j \SimpleSort{}^d_\tau(s)[i, j] \\
&= \argmax_j \left(\texttt{softmax}(-d(s_\ord{i}, s_j) / \tau)\right) \\
&= \argmin_j \left(d(s_\ord{i}, s_j)\right) \\
&= \texttt{argsort}(s)[i]
\end{align*}
as desired.
\item It suffices to show that the $i$-th row of $\SimpleSort{}^d_{\tau}(s)$ converges to the one-hot representation $h$ of $\texttt{argsort}(s)[i]$. But by part 1, the $i$-th row of $\SimpleSort{}^d_{\tau}(s)$ is the softmax of $v / \tau$ where $v$ is a vector whose unique argmax is $\texttt{argsort}(s)[i]$. Since it is a well-known property of the softmax that $\lim_{\tau \rightarrow 0^+} \texttt{softmax}(v / \tau) = h$ \cite{1994_elfadel}, we are done. 
\end{enumerate}

Note that the proof of unimodality of the \SimpleSort{} operator is straightforward, unlike the proof for the \texttt{NeuralSort} operator, which requires proving a more technical Lemma and Corollary \citep[Lemma~2, Corollary~3]{2018_grover}.

The row-stochastic property can be loosely interpreted as follows: row $r$ of \SimpleSort{} and \texttt{NeuralSort} encodes a distribution over the value of the rank $r$ element, more precisely, the probability of it being equal to $s_j$ for each $j$. In particular, note that the first row of the $\SimpleSort{}^{|\cdot|}$ operator is precisely the \texttt{softmax} of the input vector. In general, the $r$-th row of the $\SimpleSort{}^d$ operator is the \texttt{softmax} of the negative distances to the $r$-th largest element.

Finally, regarding the choice of $d$ in \SimpleSort{}, even though a large family of semi--metrics could be considered, in this work we experimented with the absolute value as well as the square distance and found the absolute value to be marginally better during experimentation. With this in consideration, in what follows we fix $d = |\cdot|$ the absolute value function, unless stated otherwise. We leave for future work learning the metric $d$ or exploring a larger family of such functions.

\section{Comparing \SimpleSort{} to \texttt{NeuralSort}}

\begin{figure*}[ht]
\begin{center}
\begin{verbatim}
def neural_sort(s, tau):
    n = tf.shape(s)[1]
    one = tf.ones((n, 1), dtype = tf.float32)
    A_s = tf.abs(s - tf.transpose(s, perm=[0, 2, 1]))
    B = tf.matmul(A_s, tf.matmul(one, tf.transpose(one)))
    scaling = tf.cast(n + 1 - 2 * (tf.range(n) + 1), dtype = tf.float32)
    C = tf.matmul(s, tf.expand_dims(scaling, 0))
    P_max = tf.transpose(C-B, perm=[0, 2, 1])
    P_hat = tf.nn.softmax(P_max / tau, -1)
    return P_hat
\end{verbatim}
\caption{Implementation of \texttt{NeuralSort} in TensorFlow as given in \cite{2018_grover}}
\label{code:neuralsort}
\end{center}
\end{figure*}

\begin{figure*}[ht]
\begin{center}
\begin{verbatim}
def soft_sort(s, tau):
    s_sorted = tf.sort(s, direction='DESCENDING', axis=1)
    pairwise_distances = -tf.abs(tf.transpose(s, perm=[0, 2, 1]) - s_sorted)
    P_hat = tf.nn.softmax(pairwise_distances / tau, -1)
    return P_hat
\end{verbatim}
\caption{Implementation of \SimpleSort{} in TensorFlow as proposed by us, with $d = |\cdot|$.}
\label{code:simplesort}
\end{center}
\vskip -0.2in
\end{figure*}

\subsection{Mathematical Simplicity}\label{sec:mathematical_simplicity}

The difference between \SimpleSort{} and \texttt{NeuralSort} becomes apparent once we write down what the actual operators look like; the equations defining them (Eq.~\ref{eq:2018_grover_eq5}, Eq.~\ref{eq:simplesort2}) are compact but do not offer much insight. Note that even though the work of \cite{2018_grover} shows that the \texttt{NeuralSort} operator has the desirable properties of Theorem~\ref{main_theorem}, the paper never gives a concrete example of what the operator instantiates to in practice, which keeps some of its complexity hidden.

Let $g_\tau : \mathbb R^{n \times n} \rightarrow \mathbb R^{n \times n}$ be the function defined as $g_\tau(X) = \texttt{softmax}(X / \tau)$, where the \texttt{softmax} is applied row-wise. Suppose that $n = 4$ and that $s$ is sorted in decreasing order $s_1 \ge s_2 \ge s_3 \ge s_4$. Then the \texttt{NeuralSort} operator is given in Figure~\ref{eq:neuralsort_logits}
and the $\SimpleSort{}$ operator is given in Figure~\ref{eq:simplesort_logits}.
Note that the diagonal of the logit matrix has been $0$-centered by subtracting a constant value from each row; this does not change the \texttt{softmax} and simplifies the expressions.
The $\SimpleSort{}$ operator is straightforward, with the $i,j$-th entry of the logit matrix given by $-|s_i - s_j|$. In contrast, the $i,j$-th entry of the \texttt{NeuralSort} operator depends on all intermediate values $s_i,s_{i+1},\dots,s_j$. This is a consequence of the coupling between the \texttt{NeuralSort} operator and the complex identity used to derive it. As we show in this paper, this complexity can be completely avoided, and results in further benefits beyond aesthetic simplicity such as flexibility, speed and mathematical simplicity.

Note that for an arbitrary region of space other than $s_1 \ge s_2 \ge s_3 \ge s_4$, the \texttt{NeuralSort} and $\SimpleSort{}$ operators look just like Figures~\ref{eq:neuralsort_logits} and~\ref{eq:simplesort_logits} respectively except for relabelling of the $s_i$ and column permutations. Indeed, we have:
\begin{proposition}\label{prop:invariance}
For both $f = \SimpleSort^d_\tau$ and $f = \texttt{NeuralSort}_\tau$, the following identity holds:
\begin{equation}
f(s) = f(\texttt{sort}(s))P_{\texttt{argsort}(s)}
\end{equation}
\end{proposition}
We defer the proof to appendix~\ref{sec:proof_of_invariance}. This proposition is interesting because it implies that the behaviour of the \SimpleSort{} and \texttt{NeuralSort} operators can be completely characterized by their functional form on the region of space where $s_1 \ge s_2 \ge \dots \ge s_n$. Indeed, for any other value of $s$, we can compute the value of $\SimpleSort(s)$ or $\texttt{NeuralSort}(s)$ by first sorting $s$, then applying $\SimpleSort$ or $\texttt{NeuralSort}$, and finally sorting the columns of the resulting matrix with the inverse permutation that sorts $s$. In particular, to our point, the proposition shows that Figures~\ref{eq:neuralsort_logits} and~\ref{eq:simplesort_logits} are valid for \textit{all} $s$ up to relabeling of the $s_i$ (by $s_\ord{i}$) and column permutations (by the inverse permutation that sorts $s$).


To further our comparison, in appendix \ref{sec:size} we show how the $\SimpleSort{}$ and \texttt{NeuralSort} operators differ in terms of the \textit{size} of their matrix entries.


\subsection{Code Simplicity}

In Figures~\ref{code:neuralsort} and~\ref{code:simplesort} we show TensorFlow implementations of the \texttt{NeuralSort} and \SimpleSort{} operators respectively. \SimpleSort{} has a simpler implementation than \texttt{NeuralSort}, which we shall see is reflected in its faster speed. (See section \ref{speed-comparison})

Note that our implementation of \SimpleSort{} is based directly off Eq.~\ref{eq:simplesort2}, and we rely on the \texttt{sort} operator. We would like to remark that there is nothing wrong with using the \texttt{sort} operator in a stochastic computation graph. Indeed, the \texttt{sort} function is continuous, almost everywhere differentiable (with non-zero gradients) and piecewise linear, just like the \texttt{max}, \texttt{min} or \texttt{ReLU} functions.

Finally, the unimodality property (Theorem~\ref{main_theorem}) implies that any algorithm that builds a relaxed permutation matrix can be used to construct the true discrete permutation matrix. This means that any relaxed sorting algorithm (in particular, \texttt{NeuralSort}) is lower bounded by the complexity of sorting, which justifies relying on sorting as a subroutine. As we show later, \SimpleSort{} is faster than \texttt{NeuralSort}. Also, we believe that this modular approach is a net positive since sorting in CPU and GPU is a well studied problem \cite{2017_singh} and any underlying improvements will benefit \SimpleSort{}'s speed as well. For instance, the current implementation in TensorFlow relies on radix sort and heap sort depending on input size.

\section{Experiments}


\begin{table*}[t]
\label{tab:run_sort}
\begin{center}
\begin{small}
\begin{sc}
\resizebox{2.0\columnwidth}{!}{
\begin{tabular}{lccccc}
\toprule
Algorithm & $n = 3$ & $n = 5$ & $n = 7$ & $n = 9$ & $n = 15$ \\
\midrule
Deterministic NeuralSort & $0.921\ \pm\ 0.006$ & $0.797\ \pm\ 0.010$ & $\textbf{0.663}\ \pm\ \textbf{0.016}$ & $\textbf{0.547}\ \pm\ \textbf{0.015}$ & $\textbf{0.253}\ \pm\ \textbf{0.021}$ \\
Stochastic NeuralSort & $0.918\ \pm\ 0.007$ & $0.801\ \pm\ 0.010$ & $\textbf{0.665}\ \pm\ \textbf{0.011}$ & $\textbf{0.540}\ \pm\ \textbf{0.019}$ & $\textbf{0.250}\ \pm\ \textbf{0.017}$ \\
\midrule
Deterministic SoftSort (Ours) & $0.918\ \pm\ 0.005$ & $0.796\ \pm\ 0.009$ & $\textbf{0.666}\ \pm\ \textbf{0.016}$ & $\textbf{0.544}\ \pm\ \textbf{0.015}$ & $\textbf{0.256}\ \pm\ \textbf{0.023}$ \\
Stochastic SoftSort (Ours) & $0.918\ \pm\ 0.005$ & $0.798\ \pm\ 0.005$ & $\textbf{0.669}\ \pm\ \textbf{0.018}$ & $\textbf{0.548}\ \pm\ \textbf{0.019}$ & $\textbf{0.250}\ \pm\ \textbf{0.020}$ \\
\midrule
\citep[reported]{2019_cuturi} & $\textbf{0.928}$ & $\textbf{0.811}$ & $\textbf{0.656}$ & $0.497$ & $0.126$ \\
\bottomrule
\end{tabular}
}
\end{sc}
\caption{Results for the sorting task averaged over 10 runs. We report the mean and standard deviation for the \textit{proportion of correct permutations}. From $n=7$ onward, the results are comparable with the state of the art.}
\end{small}
\end{center}
\end{table*}

\begin{table*}[t]
\label{tab:run_median}
\begin{center}
\begin{small}
\begin{sc}
\begin{tabular}{lccc}
\toprule
Algorithm & $n = 5$ & $n = 9$ & $n = 15$ \\
\midrule
Deterministic NeuralSort & $\textbf{21.52}\ (\textbf{0.97})$ & $\textbf{15.00}\ (\textbf{0.97})$ & $18.81\ (\textbf{0.95})$ \\
Stochastic NeuralSort & $24.78\ (\textbf{0.97})$ & $17.79\ (0.96)$ & $18.10\ (0.94)$ \\
\midrule
Deterministic SoftSort (Ours) & $23.44\ (\textbf{0.97})$ & $19.26\ (0.96)$ & $\textbf{15.54}\ (\textbf{0.95})$ \\
Stochastic SoftSort (Ours) & $26.17\ (\textbf{0.97})$ & $19.06\ (0.96)$ & $20.65\ (0.94)$ \\
\bottomrule
\end{tabular}
\end{sc}
\caption{Results for the quantile regression task. The first metric is the mean squared error ($\times 10^{-4}$) when predicting the median number. The second metric - in parenthesis - is Spearman's $R^2$ for the predictions. Results are again comparable with the state of the art.}
\end{small}
\end{center}
\end{table*}


\begin{figure*}[ht]
	\begin{center}
		\subfigure[CPU speed (in seconds) vs input dimension $n$]{\includegraphics[width=0.45\linewidth]{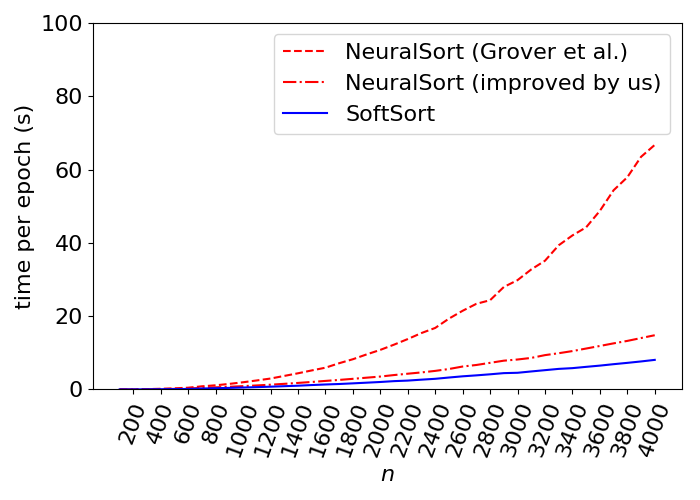}}
		\subfigure[GPU speed (in milliseconds) vs input dimension $n$]{\includegraphics[width=0.45\linewidth]{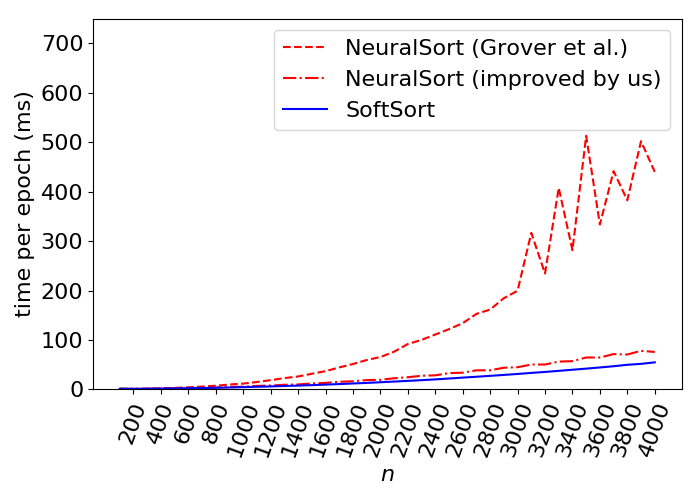}}
	\end{center}
	\caption{Speed of the \texttt{NeuralSort} and \SimpleSort{} operators on (a) CPU and (b) GPU, as a function of $n$ (the size of the vector to be sorted). Twenty vectors of size $n$ are batched together during each epoch.
	The original \texttt{NeuralSort} implementation is up to 6 times slower on both CPU and GPU. After some performance improvements, \texttt{NeuralSort} is $80\%$ slower on CPU and $40\%$ slower on GPU.}
	\label{fig:benchmark_tensorflow}
\end{figure*}

\begin{table*}[ht]
  \label{tab:knn}
  \centering
  \scriptsize
  \begin{small}
  \begin{sc}
  \begin{tabular}{|c|c|c|c|}
    \toprule
	Algorithm & MNIST & Fashion-MNIST & CIFAR-10\\
	\midrule
    kNN+Pixels & 97.2\% & 85.8\%  & 35.4\% \\
    kNN+PCA & 97.6\% & 85.9\%  & 40.9\%  \\
    kNN+AE & 97.6\% & 87.5\%  & 44.2\%\\
    \midrule
    kNN + Deterministic NeuralSort & \textbf{99.5\%} & 93.5\% & 90.7\% \\
    kNN + Stochastic NeuralSort & 99.4\%  & 93.4\% & 89.5\% \\
    \midrule
    kNN + Deterministic SoftSort (Ours) & 99.37\% & 93.60 \% & \textbf{92.03}\% \\
    kNN + Stochastic SoftSort (Ours) & 99.42\%  & \textbf{93.67}\% & 90.64\% \\
    \midrule
    CNN (w/o kNN) & 99.4\% & 93.4\% & \textbf{95.1\%} \\
        \bottomrule
  \end{tabular}
  \end{sc}
  \caption{Average test classification accuracy comparing k-nearest neighbor models. The last row includes the results from non kNN classifier. The results are comparable with the state of the art, or above it by a small margin.}
  \end{small}
\end{table*}

We first compare \SimpleSort{} to \texttt{NeuralSort} on the benchmarks from the \texttt{NeuralSort} paper \cite{2018_grover}, using the code kindly open-sourced by the authors. We show that \SimpleSort{} performs comparably to \texttt{NeuralSort}. Then, we devise a specific experiment to benchmark the speeds of the \SimpleSort{} and \texttt{NeuralSort} operators in isolation, and show that \SimpleSort{} is faster than \texttt{NeuralSort} while taking the same number of gradient steps to converge. This makes \SimpleSort{} not only the simplest, but also the fastest relaxed sorting operator proposed to date.


\subsection{Models}

For both \SimpleSort{} and \texttt{NeuralSort} we consider their deterministic and stochastic variants as in \cite{2018_grover}. The deterministic operators are those given by equations~\ref{eq:2018_grover_eq5} and~\ref{eq:simplesort2}. The stochastic variants are Plackett-Luce distributions reparameterized via Gumbel distributions \citep[Section 4.1]{2018_grover}, where the $P_{\texttt{argsort}(\cdot)}$ operator that is applied to the samples is relaxed by means of the \SimpleSort{} or \texttt{NeuralSort} operator; this is analogous to the Gumbel-Softmax trick where the \texttt{argmax} operator that is applied to the samples is relaxed via the \texttt{softmax} operator \cite{2017_jang, 2017_maddison}.

\subsection{Sorting Handwritten Numbers}

The \textit{large-MNIST} dataset of handwritten \textit{numbers} is formed by concatenating $4$ randomly selected MNIST \textit{digits}. In this task, a neural network is presented with a sequence of $n$ large-MNIST numbers and must learn the permutation that sorts these numbers. Supervision is provided only in the form of the ground-truth permutation. Performance on the task is measured by:
\begin{enumerate}
    \item The proportion of \textit{permutations} that are perfectly recovered.
    \item The proportion of \textit{permutation elements} that are correctly recovered.
\end{enumerate}
Note that the first metric is always less than or equal to the second metric. We use the setup in \cite{2019_cuturi} to be able to compare against their Optimal-Transport-based method. They use $100$ epochs to train all models.

The results for the first metric are shown in Table~\ref{tab:run_sort}. We report the mean and standard deviation over 10 runs. We see that \SimpleSort{} and \texttt{NeuralSort} perform identically for all values of $n$. Moreover, our results for \texttt{NeuralSort} are better than those reported in \cite{2019_cuturi}, to the extent that \texttt{NeuralSort} and \SimpleSort{} outperform the method of \cite{2019_cuturi} for $n = 9, 15$, unlike reported in said paper. We found that the hyperparameter values reported in \cite{2018_grover} and used by \cite{2019_cuturi} for \texttt{NeuralSort} are far from ideal: \cite{2018_grover} reports using a learning rate of $10^{-4}$ and temperature values from the set $\{1,2,4,8,16\}$. However, a higher learning rate dramatically improves \texttt{NeuralSort}'s results, and higher temperatures also help. Concretely, we used a learning rate of $0.005$ for all the \SimpleSort{} and \texttt{NeuralSort} models, and a value of $\tau = 1024$ for $n = 3,5,7$ and $\tau = 128$ for $n = 9, 15$. The results for the second metric are reported in appendix \ref{app:run_sort_aux}. In the appendix we also include learning curves for \SimpleSort{} and \texttt{NeuralSort}, which show that they converge at the same speed.


\subsection{Quantile Regression}

As in the sorting task, a neural network is presented with a sequence of $n$ large-MNIST numbers. The task is to predict the median element from the sequence, and this is the only available form of supervision. Performance on the task is measured by mean squared error and Spearman's rank correlation. We used $100$ iterations to train all models.

The results are shown in Table~\ref{tab:run_median}. We used a learning rate of $10^{-3}$ for all models - again, higher than that reported in \cite{2018_grover} - and grid-searched the temperature on the set $\{128, 256, 512, 1024, 2048, 4096\}$ - again, higher than that reported in \cite{2018_grover}. We see that \SimpleSort{} and \texttt{NeuralSort} perform similarly, with \texttt{NeuralSort} sometimes outperforming \SimpleSort{} and vice versa. The results for \texttt{NeuralSort} are also much better than those reported in \cite{2018_grover}, which we attribute to the better choice of hyperparameters, concretely, the higher learning rate. In the appendix we also include learning curves for \SimpleSort{} and \texttt{NeuralSort}, which show that they converge at the same speed.

\subsection{Differentiable kNN}

In this setup, we explored using the \SimpleSort{} operator to learn a differentiable $k$-nearest neighbours (kNN) classifier that is able to learn a representation function, used to measure the distance between the candidates.

In a supervised training framework, we have a dataset that consists of pairs $(x, y)$ of a datapoint and a label. We are interested in learning a map $\Phi$ to embed every $x$ such that we can use a kNN classifier to identify the class of $\hat{x}$ by looking at the class of its closest neighbours according to the distance $\|\Phi(x) - \Phi(\hat{x})\|$. Such a classifier would be valuable by virtue of being interpretable and robust to both noise and unseen classes.

This is achieved by constructing episodes during training that consist of one pair $\hat{x}, \hat{y}$ and $n$ candidate pairs $(x_i, y_i)$ for $i = 1\dots n$, arranged in two column vectors $X$ and $Y$. The probability $P(\hat{y}|\hat{x},X,Y)$ of class $\hat{y}$ under a kNN classifier is the average of the first $k$ entries in the vector
\[
P_{\texttt{argsort}(-\|\Phi(X) - \Phi(\hat{x})\|^2)} \mathbb{I}_{Y = \hat{y}}
\]
where $\|\Phi(X) - \Phi(\hat{x})\|^2$ is the vector of squared distances from the candidate points and $\mathbb{I}_{Y = \hat{y}}$ is the binary vector indicating which indexes have class $\hat{y}$.
Thus, if we replace $P_\texttt{argsort}$ by the \SimpleSort{} operator we obtain a differentiable relaxation $\widehat{P}(\hat{y}|\hat{x},X,Y)$. To compute the loss we follow \cite{2018_grover} and use $-\widehat{P}(\hat{y}|\hat{x},X,Y)$. We also experimented with the cross entropy loss, but the performance went down for both methods.

When $k=1$, our method is closely related to \emph{Matching Networks} \cite{2016_vinyals}. This follows from the following result: (See proof in Appendix~\ref{sec:proof_of_dknn})
\begin{proposition}\label{dknn}
Let $k=1$ and $\widehat{P}$ be the differentiable kNN operator using $\SimpleSort{}^{|\cdot|}_2$. If we choose the embedding function $\Phi$ to be of norm $1$, then
\begin{equation*}
    \widehat{P}(\hat{y}|\hat{x},X,Y) =
    {\sum_{i:y_i = \hat{y}} e^{\Phi(\hat{x})\cdot\Phi(x_i)}} \bigg/
    {\sum_{i=1\dots n} e^{\Phi(\hat{x})\cdot\Phi(x_i)}}
\end{equation*}
\end{proposition}
This suggests that our method is a generalization of \emph{Matching Networks}, since in our experiments larger values of $k$ yielded better results consistently and we expect a kNN classifier to be more robust in a setup with noisy labels. However, \emph{Matching Networks} use contextual embedding functions, and different networks for $\hat{x}$ and the candidates $x_i$, both of which could be incorporated to our setup. A more comprehensive study comparing both methods on a few shot classification dataset such as \emph{Omniglot}~\cite{2011_lake} is left for future work.

We applied this method to three benchmark datasets: MNIST, Fashion MNIST and CIFAR-10 with canonical splits. As baselines, we compare against \texttt{NeuralSort} as well as other kNN models with fixed representations coming from raw pixels, a PCA feature extractor and an auto-encoder. All the results are based on the ones reported in \cite{2018_grover}. We also included for comparison a standard classifier using a convolutional network.

Results are shown in Table \ref{tab:knn}. In every case, we achieve comparable accuracies with \texttt{NeuralSort} implementation, either slightly outperforming or underperforming \texttt{NeuralSort}. See hyperparameters used in appendix \ref{sec:experimental_details}.

\subsection{Speed Comparison}
\label{speed-comparison}

We set up an experiment to compare the speed of the \SimpleSort{} and \texttt{NeuralSort} operators. We are interested in exercising both their forward and backward calls. To this end, we set up a dummy learning task where the goal is to perturb an $n$-dimensional input vector $\theta$ to make it become sorted. We scale $\theta$ to $[0, 1]$ and feed it through the \SimpleSort{} or \texttt{NeuralSort} operator to obtain $\widehat{P}(\theta)$, and place a loss on $\widehat{P}(\theta)$ that encourages it to become equal to the identity matrix, and thus encourages the input to become sorted.


Concretely, we place the cross-entropy loss between the true permutation matrix and the predicted soft URS matrix:
\begin{equation*}
L(\widehat{P}) = - \frac{1}{n}\sum_{i = 1}^n \log \widehat{P}[i, i]
\end{equation*}
This encourages the probability mass from each row of $\widehat{P}$ to concentrate on the diagonal, which drives $\theta$ to sort itself. Note that this is a trivial task, since for example a pointwise ranking loss $\frac{1}{n} \sum_{i = 1}^n (\theta_i + i)^2$ \citep[Section~2.2]{2008_taylor} leads the input to become sorted too, without any need for the \SimpleSort{} or \texttt{NeuralSort} operators. However, this task is a reasonable benchmark to measure the speed of the two operators in a realistic learning setting.

We benchmark $n$ in the range $100$ to $4000$, and batch $20$ inputs $\theta$ together to exercise batching. Thus the input is a parameter tensor of shape $20 \times n$. Models are trained for $100$ epochs, which we verified is enough for the parameter vectors to become perfectly sorted by the end of training (i.e., to succeed at the task).

In Figure~\ref{fig:benchmark_tensorflow} we show the results for the TensorFlow implementations of \texttt{NeuralSort} and \SimpleSort{} given in Figures~\ref{code:neuralsort} and~\ref{code:simplesort} respectively. We see that on both CPU and GPU, \SimpleSort{} is faster than \texttt{NeuralSort}. For $N = 4000$, \SimpleSort{} is about 6 times faster than the \texttt{NeuralSort} implementation of~\cite{2018_grover} on both CPU and GPU. We tried to speed up the \texttt{NeuralSort} implementation of~\cite{2018_grover}, and although we were able to improve it, \texttt{NeuralSort} was still slower than \SimpleSort{}, concretely: $80\%$ slower on CPU and $40\%$ slower on GPU. Details of our improvements to the speed of the \texttt{NeuralSort} operator are provided in appendix~\ref{appendix:neuralsort_performance_improvement}.

The performance results for PyTorch are provided in the appendix and are similar to the TensorFlow results. In the appendix we also show that the learning curves of \SimpleSort{} with $d = |\cdot|^2$ and \texttt{NeuralSort} are almost identical; interestingly, we found that using $d = |\cdot|$ converges more slowly on this synthetic task.

We also investigated if relying on a sorting routine could cause slower run times in worst-case scenarios. When using sequences sorted in the opposite order we did not note any significant slowdowns. Furthermore, in applications where this could be a concern, the effect can be avoided entirely by shuffling the inputs before applying our operator.

As a final note, given that the cost of sorting is sub-quadratic, and most of the computation is payed when building and applying the $n \times n$ matrix, we also think that our algorithm could be made faster asymptotically by constructing sparse versions of the \SimpleSort{} operator. For applications like differentiable nearest neighbors, evidence suggests than processing longer sequences yields better results, which motivates improvements in the asymptotic complexity.
We leave this topic for future work.

\section{Conclusion}

We have introduced \SimpleSort{}, a simple continuous relaxation for the \texttt{argsort} operator. The $r$-th row of the \SimpleSort{} operator is simply the \texttt{softmax} of the negative distances to the $r$-th largest element.

\SimpleSort{} has similar properties to those of the \texttt{NeuralSort} operator of \cite{2018_grover}. However, due to its simplicity, \SimpleSort{} is trivial to implement, more modular, faster than \texttt{NeuralSort}, and proofs regarding the \SimpleSort{} operator are effortless. We also empirically find that it is just as easy to optimize.

Fundamentally, \SimpleSort{} advances the state of the art in differentiable sorting by significantly simplifying previous approaches.
Our code and experiments can be found at \url{https://github.com/sprillo/softsort}.

\clearpage

\section*{Acknowledgments}
We thank Assistant Professor Jordan Boyd-Graber from University of Maryland and Visting Researcher at Google, and Thomas Müller from Google Language Research at Zurich for their feedback and comments on earlier versions of the manuscript. We would also like to thank the anonymous reviewers for their feedback that helped improve this work.

\bibliography{references}
\bibliographystyle{icml2020}

\clearpage\input{sections/appendix}

\end{document}

%% file: preamble.tex
\usepackage{microtype}
\usepackage{graphicx}
\usepackage{subfigure}
\usepackage{booktabs} 

\usepackage{hyperref}
\usepackage{amsfonts}
\usepackage{amsmath}
\usepackage{amssymb}
\usepackage{dsfont}
\usepackage{textcomp}
\usepackage{tikz}
\usetikzlibrary{positioning}

\newtheorem{theorem}{Theorem}
\newtheorem{proposition}{Proposition}

\newtheorem{lemma}{Lemma}
\newtheorem{definition}{Definition}

\DeclareMathOperator*{\argmax}{arg\,max}
\DeclareMathOperator*{\argmin}{arg\,min}
\newcommand{\SimpleSort}{\texttt{SoftSort}}
\newcommand\ord[1]{{[#1]}}

\newcommand{\note}[4][]{} %

%% file: sections/appendix.tex
\appendix

\section{Experimental Details}\label{sec:experimental_details}

We use the code kindly open sourced by \cite{2018_grover} to perform the sorting, quantile regression, and kNN experiments. As such, we are using the same setup as in \cite{2018_grover}. The work of \cite{2019_cuturi} also uses this code for the sorting task, allowing for a fair comparison.

To make our work self-contained, in this section we recall the main experimental details from \cite{2018_grover}, and we also provide our hyperparameter settings, which crucially differ from those used in \cite{2018_grover} by the use of higher learning rates and temperatures (leading to improved results).

\subsection{Sorting Handwritten Numbers}

\begin{figure*}[ht]
	\begin{center}
		\subfigure[Sorting handwritten numbers learning curves.]{\includegraphics[width=0.45\linewidth]{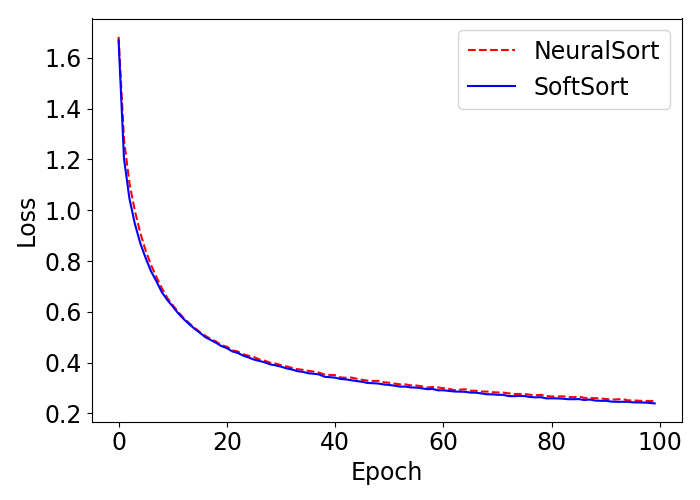}}
		\subfigure[Quantile regression learning curves]{\includegraphics[width=0.45\linewidth]{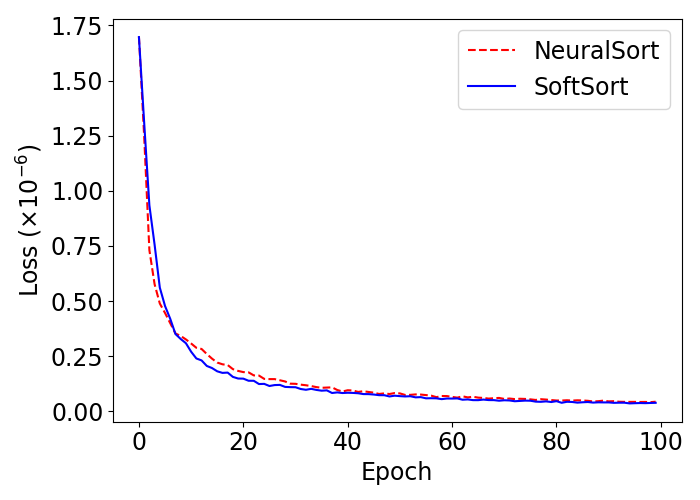}}
	\end{center}
	\caption{Learning curves for the `sorting handwritten numbers' and `quantile regression' tasks. The learning curves for \SimpleSort{} and \texttt{NeuralSort} are almost identical.}
	\label{fig:learning_curves_sort_and_median}
\end{figure*}

\subsubsection{Architecture}

The convolutional neural network architecture used to map $112\times 28$ large-MNIST images to scores is as follows:

\begin{align*}
& \text{Conv[Kernel: 5x5, Stride: 1, Output: 112x28x32, } \\
& \quad \quad\ \ \text{Activation: Relu]} \\
\rightarrow & \text{Pool[Stride: 2, Output: 56x14x32]} \\
\rightarrow & \text{Conv[Kernel: 5x5, Stride: 1, Output: 56x14x64, } \\
& \quad \quad\ \ \text{Activation: Relu]} \\
\rightarrow & \text{Pool[Stride: 2, Output: 28x7x64]} \\
\rightarrow & \text{FC[Units: 64, Activation: Relu]} \\
\rightarrow & \text{FC[Units: 1, Activation: None]}
\end{align*}

Recall that the large-MNIST dataset is formed by concatenating four $28\times 28$ MNIST images, hence each large-MNIST image input is of size $112 \times 28$.

For a given input sequence $x$ of large-MNIST images, using the above CNN we obtain a vector $s$ of scores (one score per image). Feeding this score vector into \texttt{NeuralSort} or \SimpleSort{} yields the matrix $\widehat P(s)$ which is a relaxation for $P_{\texttt{argsort}(s)}$.

\subsubsection{Loss Functions}

To lean to sort the input sequence $x$ of large-MNIST digits, \cite{2018_grover} imposes a cross-entropy loss between the rows of the true permutation matrix $P$ and the learnt relaxation $\widehat P(s)$, namely:
\[L = \frac{1}{n} \sum_{i, j = 1}^n \mathds 1\{P[i, j] = 1\}\log \widehat P(s)[i, j]\]
This is the loss for one example $(x, P)$ in the deterministic setup. For the stochastic setup with reparameterized Plackett-Luce distributions, the loss is instead:
\[L = \frac{1}{n} \sum_{i, j = 1}^n \sum_{k = 1}^{n_s} \mathds 1\{P[i, j] = 1\}\log \widehat P(s + z_k)[i, j]\]
where $z_k$ $(1 \le k \le n_s)$ are samples from the Gumbel distribution.

\subsubsection{Hyperparamaters}

For this task we used an Adam optimizer with an initial learning rate of $0.005$ and a batch size of $20$. The temperature $\tau$ was selected from the set $\{1, 16, 128, 1024\}$ based on validation set accuracy on predicting entire permutations. As a results, we used a value of $\tau = 1024$ for $n = 3,5,7$ and $\tau = 128$ for $n = 9, 15$. $100$ iterations were used to train all models. For the stochastic setting, $n_s = 5$ samples were used.

\subsubsection{Learning Curves}

In Figure~\ref{fig:learning_curves_sort_and_median} (a) we show the learning curves for deterministic \SimpleSort{} and \texttt{NeuralSort}, with $N = 15$. These are the average learning curves over all $10$ runs. Both learning curves are almost identical, showing that in this task both operators can essentially be exchanged for one another.

\subsection{Quantile Regression}

\subsubsection{Architecture}

The same convolutional neural network as in the sorting task was used to map large-MNIST images to scores. A second neural network $g_{\theta}$ with the same architecture but different parameters is used to regress each image to its value. These two networks are then used by the loss functions below to learn the median element.

\subsubsection{Loss Functions}

To learn the median element of the input sequence $x$ of large-MNIST digits, \cite{2018_grover} first soft-sorts $x$ via $\widehat P(s)x$ which allows extracting the candidate median image.
This candidate median image is then mapped to its predicted value $\hat y$ via the CNN $g_{\theta}$. The square loss between $\hat y$ and the true median value $y$ is incurred.
As in \citep[Section E.2]{2018_grover}, the loss for a single example $(x, y)$ can thus compactly be written as (in the deterministic case):
\[L = \|y - g_{\theta}(\widehat P(s)x)\|_2^2\]
For the stochastic case, the loss for a single example $(x, y)$ is instead:
\[L = \sum_{k = 1}^{n_s} \|y - g_{\theta}(\widehat P(s + z_k)x)\|_2^2\]
where $z_k$ $(1 \le k \le n_s)$ are samples from the Gumbel distribution.

\subsubsection{Hyperparamaters}

We used an Adam optimizer with an initial learning rate of $0.001$ and a batch size of $5$. The value of $\tau$ was grid searched on the set $\{128, 256, 512, 1024, 2048, 4096\}$ based on validation set MSE. The final values of $\tau$ used to train the models and evaluate test set performance are given in Table~\ref{tab:run_median_taus}. $100$ iterations were used to train all models. For the stochastic setting, $n_s = 5$ samples were used.

\subsubsection{Learning Curves}

In Figure~\ref{fig:learning_curves_sort_and_median} (b) we show the learning curves for deterministic \SimpleSort{} and \texttt{NeuralSort}, with $N = 15$. Both learning curves are almost identical, showing that in this task both operators can essentially be exchanged for one another.

\begin{table}[t]
\caption{Values of $\tau$ used for the quantile regression task.}
\label{tab:run_median_taus}
\vskip 0.15in
\begin{center}
\begin{small}
\begin{sc}
\begin{tabular}{lccc}
\toprule
Algorithm & $n = 5$ & $n = 9$ & $n = 15$ \\
\midrule
Deterministic NeuralSort & 1024 & 512 & 1024 \\
Stochastic NeuralSort & 2048 & 512 & 4096 \\
\midrule
Deterministic SoftSort & 2048 & 2048 & 256 \\
Stochastic SoftSort & 4096 & 2048 & 2048 \\
\bottomrule
\end{tabular}
\end{sc}
\end{small}
\end{center}
\vskip -0.1in
\end{table}

\subsection{Differentiable KNN}

\subsubsection{Architectures}

To embed the images before applying differentiable kNN, we used the following convolutional network architectures. For MNIST:
\begin{align*}
& \text{Conv[Kernel: 5x5, Stride: 1, Output: 24x24x20, } \\
& \quad \quad\ \ \text{Activation: Relu]} \\
\rightarrow & \text{Pool[Stride: 2, Output: 12x12x20]} \\
\rightarrow & \text{Conv[Kernel: 5x5, Stride: 1, Output: 8x8x50, } \\
& \quad \quad\ \ \text{Activation: Relu]} \\
\rightarrow & \text{Pool[Stride: 2, Output: 4x4x50]} \\
\rightarrow & \text{FC[Units: 50, Activation: Relu]}
\end{align*}
and for Fashion-MNIST and CIFAR-10 we used the \emph{ResNet18} architecture \cite{He2016IdentityMI} as defined in \hyperlink{https://github.com/kuangliu/pytorch-cifar}{github.com/kuangliu/pytorch-cifar}, but we keep the $512$ dimensional output before the last classification layer.

For the baseline experiments in pixel distance with PCA and kNN, we report the results of \cite{2018_grover}, using \textit{scikit-learn} implementations.

In the auto-encoder baselines, the embeddings were trained using the follow architectures. For MNIST and Fashion-MNIST:
\begin{align*}
            & \text{Encoder:} \\
            & \text{FC[Units: 500, Activation: Relu]} \\
\rightarrow & \text{FC[Units: 500, Activation: Relu]} \\
\rightarrow & \text{FC[Units: 50, Activation: Relu]} \\
            & \text{Decoder:} \\
\rightarrow & \text{FC[Units: 500, Activation: Relu]} \\
\rightarrow & \text{FC[Units: 500, Activation: Relu]} \\
\rightarrow & \text{FC[Units: 784, Activation: Sigmoid]}
\end{align*}

For CIFAR-10, we follow the architecture defined at
\hyperlink{https://github.com/shibuiwilliam/Keras_Autoencoder}{github.com/shibuiwilliam/Keras\_Autoencoder}, with a bottleneck dimension of $256$.

\subsubsection{Loss Functions}

For the models using \SimpleSort{} or \texttt{NeuralSort} we use the negative of the probability output from the kNN model as a loss function. 
For the auto-encoder baselines we use a per-pixel binary cross entropy loss.

\subsubsection{Hyperparamaters}

We perform a grid search for $k \in (1, 3, 5, 9)$, 
$\tau\in(1, 4, 16, 64, 128, 512)$, learning rates taking values in 
$10^{-3}$, $10^{-4}$ and $10^{-5}$. We train for $200$ epochs and choose the model based on validation loss. The optimizer used is \emph{SGD} with momentum of $0.9$. Every batch has $100$ episode, each containing $100$ candidates.

\subsection{Speed Comparison}\label{sec:speed_comparison_appendix}

\subsubsection{Architecture}

The input parameter vector $\theta$ of shape $20 \times n$ (20 being the batch size) is first normalized to $[0, 1]$ and then fed through the \texttt{NeuralSort} or \SimpleSort{} operator, producing an output tensor $\widehat P$ of shape $20 \times n \times n$.

\subsubsection{Loss Function}

We impose the following loss term over the batch:
\[L(\widehat P) = - \frac{1}{20} \sum_{i = 1}^{20} \frac{1}{n} \sum_{j = 1}^n \log \widehat P[i, j, j]\]
This loss term encourages the probability mass from each row of $\widehat P[i, :, :]$ to concentrate on the diagonal, i.e. encourages each row of $\theta$ to become sorted in decreasing order. We also add an $L2$ penalty term $\frac{1}{200}\|\theta\|_2^2$ which ensures that the entries of $\theta$ do not diverge during training. 

\subsubsection{Hyperparamaters}

We used $100$ epochs to train the models, with the first epoch used as burn-in to warm up the CPU or GPU (i.e. the first epoch is excluded from the time measurement). We used a temperature of $\tau = 100.0$ for \texttt{NeuralSort} and $\tau = 0.03, d = |\cdot|^2$ for \SimpleSort{}. The entries of $\theta$ are initialized uniformly at random in $[-1, 1]$. A momentum optimizer with learning rate $10$ and momentum $0.5$ was used. With these settings, 100 epochs are enough to sort each row of $\theta$ in decreasing order perfectly for $n = 4000$.

Note that since the goal is to benchmark the operator's speeds, performance on the Spearman rank correlation metric is anecdotal. However, we took the trouble of tuning the hyperparameters and the optimizer to make the learning setting as realistic as possible, and to ensure that the entries in $\theta$ are not diverging (which would negatively impact and confound the performance results). Finally, note that the learning problem is trivial, as a pointwise loss such as $\sum_{i = 1}^{20}\sum_{j = 1}^n (\theta_{ij} + j)^2$ sorts the rows of $\theta$ without need for the \texttt{NeuralSort} or \SimpleSort{} operator. However, this bare-bones task exposes the computational performance of the \texttt{NeuralSort} and \SimpleSort{} operators.

\subsubsection{Learning Curves}

\begin{figure}[ht]
\begin{center}
\centerline{\includegraphics[width=\columnwidth]{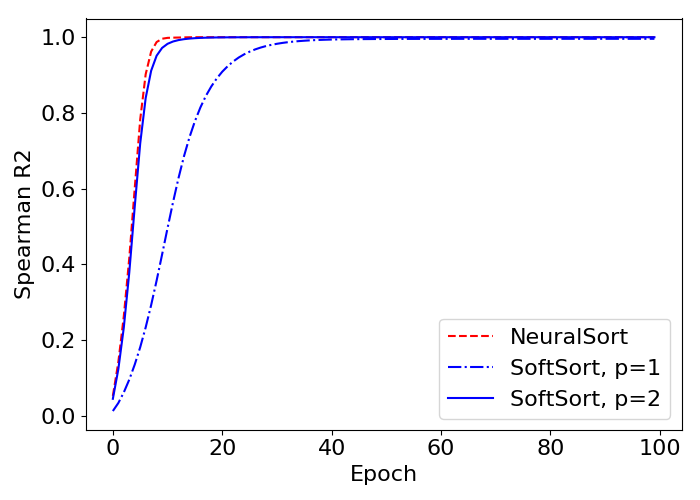}}
\caption{Learning curves for \SimpleSort{} with $d = |\cdot|^p$ for $p \in \{1, 2\}$, and \texttt{NeuralSort}, on the speed comparison task.}
\label{fig:learning_curves_synthetic}
\end{center}
\end{figure}

In Figure~\ref{fig:learning_curves_synthetic} we show the learning curves for $N = 4000$; the Spearman correlation metric is plotted against epoch. We see that \SimpleSort{} with $d = |\cdot|^2$ and \texttt{NeuralSort} have almost identical learning curves. Interestingly, \SimpleSort{} with $d = |\cdot|$ converges more slowly.


\subsubsection{NeuralSort Performance Improvement}\label{appendix:neuralsort_performance_improvement}

We found that the \texttt{NeuralSort} implementations provided by ~\cite{2018_grover} in both TensorFlow and PyTorch have complexity $\mathcal O(n^3)$. Indeed, in their TensorFlow implementation (Figure~\ref{code:neuralsort}), the complexity of the following line is $\mathcal O(n^3)$:
\begin{verbatim}
B = tf.matmul(A_s, tf.matmul(one,
              tf.transpose(one)))
\end{verbatim}
since the three matrices multiplied have sizes $n \times n$, $n \times 1$, and $1 \times n$ respectively. To obtain $\mathcal O(n^2)$ complexity we associate differently:
\begin{verbatim}
B = tf.matmul(tf.matmul(A_s, one),
              tf.transpose(one))
\end{verbatim}
The same is true for their PyTorch implementation (Figure~\ref{code:neuralsort_pytorch}). This way, we were able to speed up the implementations provided by ~\cite{2018_grover}.

\subsubsection{PyTorch Results}

\begin{figure*}[ht]
	\begin{center}
		\subfigure[CPU speed vs input dimension $n$]{\includegraphics[width=0.45\linewidth]{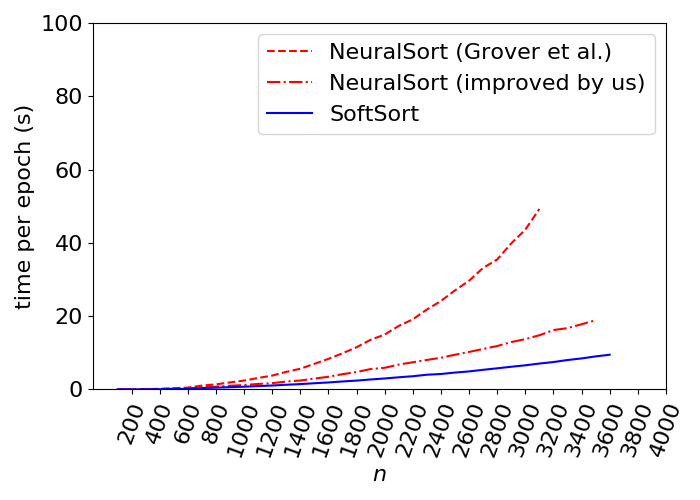}}
		\subfigure[GPU speed vs input dimension $n$]{\includegraphics[width=0.45\linewidth]{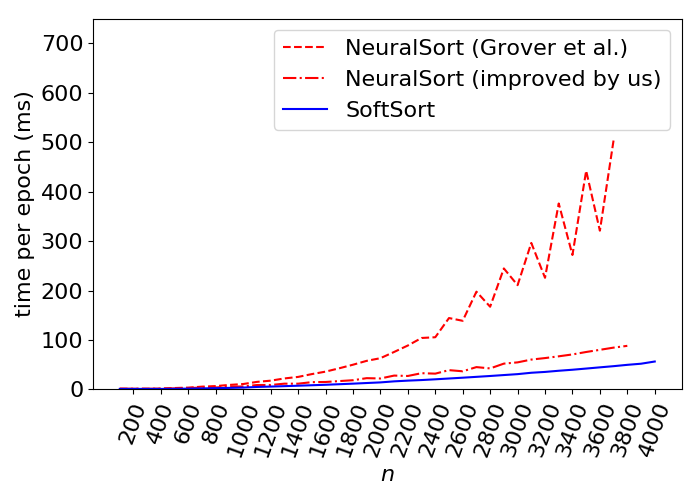}}
	\end{center}
	\caption{Speed of the \texttt{NeuralSort} and \SimpleSort{} operators on (a) CPU and (b) GPU, as a function of $n$ (the size of the vector to be sorted). Twenty vectors of size $n$ are batched together during each epoch. Note that CPU plot y-axis is in seconds (s), while GPU plot y-axis is in milliseconds (ms). Implementation in PyTorch.}
	\label{fig:benchmark_pytorch}
\end{figure*}

In Figure~\ref{fig:benchmark_pytorch} we show the benchmarking results for the PyTorch framework \cite{2017_paszke}. These are analogous to the results presented in Figure~\ref{fig:benchmark_tensorflow}) of the main text. The results are similar to those for the TensorFlow framework, except that for PyTorch, \texttt{NeuralSort} runs out of memory on CPU for $n = 3600$, on GPU for $n = 3900$, and \SimpleSort{} runs out of memory on CPU for $n = 3700$.

\subsubsection{Hardware Specification}


We used a GPU V100 and an n1-highmem-2 (2 vCPUs, 13 GB memory) Google Cloud instance to perform the speed comparison experiment.

We were also able to closely reproduce the GPU results on an Amazon EC2 p2.xlarge instance (4 vCPUs, 61 GB memory) equipped with a GPU Tesla K80, and the CPU results on an Amazon EC2 c5.2xlarge instance (8 vCPUs, 16 GB memory).

\section{Proof of Proposition~\ref{prop:invariance}}\label{sec:proof_of_invariance}

First we recall the Proposition:

\textbf{Proposition}. For both $f = \SimpleSort^d_\tau$ (with any $d$) and $f = \texttt{NeuralSort}_\tau$, the following identity holds:
\begin{equation}
f(s) = f(\texttt{sort}(s))P_{\texttt{argsort}(s)}
\end{equation}

To prove the proposition, we will use the following two Lemmas:

\begin{lemma}\label{lemma:mult_by_P}
Let $P \in \mathbb R^{n \times n}$ be a permutation matrix, and let $g : \mathbb R^k \rightarrow \mathbb R$ be any function. Let $G : \underbrace{\mathbb R^{n \times n} \times \dots \times \mathbb R^{n \times n}}_{k\text{ times}} \rightarrow \mathbb R^{n \times n}$ be the pointwise application of $g$, that is:
\begin{equation}\label{eq:pointwise_def}
G(A_1,\dots,A_k)_{i,j} = g((A_1)_{i,j},\dots,(A_k)_{i,j})
\end{equation}
Then the following identity holds for any $A_1,\dots,A_k \in \mathbb R^{n \times n}$:
\begin{equation}
G(A_1,\dots,A_k)P = G(A_1P,\dots,A_kP)
\end{equation}
\end{lemma}

\textbf{Proof of Lemma~\ref{lemma:mult_by_P}}. Since $P$ is a permutation matrix, multiplication to the right by $P$ permutes columns according to some permutation, i.e.
\begin{equation}\label{eq:A_times_P}
(AP)_{i,j} = A_{i,\pi(j)}
\end{equation}
for some permutation $\pi$ and any $A \in \mathbb R^{n \times n}$. Thus, for any fixed $i,j$:
\begin{align*}
& (G(A_1,\dots,A_k)P)_{i,j}\\
\stackrel{(i)}{=} & G(A_1,\dots,A_k)_{i,\pi(j)} \\
\stackrel{(ii)}{=} & g((A_1)_{i,\pi(j)}, \dots, (A_k)_{i, \pi(j)}) \\
\stackrel{(iii)}{=} & g((A_1P)_{i,j},\dots,(A_kP)_{i,j}) \\
\stackrel{(iv)}{=} & G(A_1P,\dots,A_kP)_{i,j}
\end{align*}
where $(i), (iii)$ follow from Eq.~\ref{eq:A_times_P}, and $(ii), (iv)$ follow from Eq.~\ref{eq:pointwise_def}. This proves the Lemma. $\blacksquare$

\begin{lemma}\label{lemma:softmax_and_P}
Let $P \in \mathbb R^{n \times n}$ be a permutation matrix, and $\sigma = \texttt{softmax}$ denote the row-wise softmax, i.e.:
\begin{equation}\label{eq:row_wise_softmax_def}
\sigma(A)_{i,j} = \frac{\exp\{A_{i, j}\}}{\sum_{k} \exp\{A_{i, k}\}}
\end{equation}
Then the following identity holds for any $A \in \mathbb R^{n \times n}$:
\begin{equation}\label{eq:softmax_and_P}
\sigma(A)P = \sigma(AP)
\end{equation}
\end{lemma}

\textbf{Proof of Lemma~\ref{lemma:softmax_and_P}}. As before, there exists a permutation $\pi$ such that:
\begin{equation}\label{B_times_P}
(BP)_{i,j} = B_{i, \pi(j)}
\end{equation} for any $B \in \mathbb R^{n \times n}$. Thus for any fixed $i,j$:
\begin{align*}
& (\sigma(A)P)_{i,j} \\
\stackrel{(i)}{=} & \sigma(A)_{i,\pi(j)} \\
\stackrel{(ii)}= & \frac{\exp\{A_{i, \pi(j)}\}}{\sum_k \exp\{A_{i, \pi(k)}\}} \\
\stackrel{(iii)}= & \frac{\exp\{(AP)_{i,j}\}}{\sum_k \exp\{(AP)_{i,k}\}} \\
\stackrel{(iv)}= & \sigma(AP)_{i,j}
\end{align*}
where $(i), (iii)$ follow from Eq.~\ref{B_times_P} and $(ii),(iv)$ follow from the definition of the row-wise softmax (Eq.~\ref{eq:row_wise_softmax_def}). This proves the Lemma. $\blacksquare$



We now leverage the Lemmas to provide proofs of Proposition~\ref{prop:invariance} for each operator. To unclutter equations, we will denote by $\sigma = \texttt{softmax}$ the row-wise softmax operator.

\textbf{Proof of Proposition~\ref{prop:invariance} for \SimpleSort}. We have that:
\begin{align*}
& \SimpleSort^d_\tau(\texttt{sort}(s))P_{\texttt{argsort}(s)}\\
\stackrel{(i)}{=} & \sigma\Big(\frac{-d(\texttt{sort}(\texttt{sort}(s)) \mathds 1^T, \mathds 1 \texttt{sort}(s)^T)}{\tau}\Big)P_{\texttt{argsort}(s)} \\
\stackrel{(ii)}{=} & \sigma\Big(\frac{-d(\texttt{sort}(s) \mathds 1^T, \mathds 1 \texttt{sort}(s)^T)}{\tau}\Big)P_{\texttt{argsort}(s)}
\end{align*}
where $(i)$ follows from the definition of the \SimpleSort{} operator (Eq.~\ref{eq:simplesort2}) and $(ii)$ follows from the idempotence of the \texttt{sort} operator, i.e. $\texttt{sort}(\texttt{sort}(s)) = \texttt{sort}(s)$. Invoking Lemma~\ref{lemma:softmax_and_P}, we can push $P_{\texttt{argsort}(s)}$ into the softmax:
\begin{align*}
= & \sigma\Big(\frac{-d(\texttt{sort}(s) \mathds 1^T, \mathds 1 \texttt{sort}(s)^T)}{\tau}P_{\texttt{argsort}(s)}\Big)
\end{align*}
Using Lemma~\ref{lemma:mult_by_P} we can further push $P_{\texttt{argsort}(s)}$ into the pointwise $d$ function:
\begin{align*}
= & \sigma\Big(\frac{-d(\texttt{sort}(s) \mathds 1^T P_{\texttt{argsort}(s)}, \mathds 1 \texttt{sort}(s)^T P_{\texttt{argsort}(s)})}{\tau}\Big)
\end{align*}
Now note that $\mathds 1^T P_{\texttt{argsort}(s)} = \mathds 1^T$ since $P$ is a permutation matrix and thus the columns of $P$ add up to $1$. Also, since $\texttt{sort}(s)^T = P_{\texttt{argsort}(s)}s$ then $\texttt{sort}(s)^T P_{\texttt{argsort}(s)} = s^T P_{\texttt{argsort}(s)}^T P_{\texttt{argsort}(s)} = s^T$ since $P_{\texttt{argsort}(s)}^T P_{\texttt{argsort}(s)} = I$ (because $P_{\texttt{argsort}(s)}$ is a permutation matrix). Hence we arrive at:
\begin{align*}
= & \sigma\Big(\frac{-d(\texttt{sort}(s) \mathds 1^T, \mathds 1 s^T)}{\tau}\Big) \\
= & \SimpleSort^d_\tau(s)
\end{align*}
which proves the proposition for \SimpleSort{}. $\blacksquare$

\textbf{Proof of Proposition~\ref{prop:invariance} for \texttt{NeuralSort}}. For any fixed $i$, inspecting row $i$ we get:
\begin{align*}
& (\texttt{NeuralSort}_\tau(\texttt{sort}(s))P_{\texttt{argsort}(s)})[i, :] \\
\stackrel{(i)}{=} & (\texttt{NeuralSort}_\tau(\texttt{sort}(s))[i, :])P_{\texttt{argsort}(s)} \\
\stackrel{(ii)}{=} & \sigma\Big(\frac{(n + 1 - 2i)\texttt{sort}(s)^T - \mathds 1^T A_{\texttt{sort}(s)}^T}{\tau}\Big) P_{\texttt{argsort}(s)}
\end{align*}

where $(i)$ follows since row-indexing and column permutation trivially commute, i.e. $(BP)[i, :] = (B[i, :])P$ for any $B \in \mathbb R^{n \times n}$, and $(ii)$ is just the definition of \texttt{NeuralSort} (Eq.~\ref{eq:2018_grover_eq5}, taken as a row vector).

Using Lemma~\ref{lemma:softmax_and_P} we can push $P_{\texttt{argsort}(s)}$ into the softmax, and so we get:
\begin{align}\label{eq:choclo}
= \sigma\Big((& (n + 1 - 2i)\texttt{sort}(s)^T P_{\texttt{argsort}(s)} \nonumber \\
& - \mathds 1^T A_{\texttt{sort}(s)}^T P_{\texttt{argsort}(s)})/\tau\Big)
\end{align}
Now note that $\texttt{sort}(s)^TP_{\texttt{argsort}(s)} = s^T$ (as we showed in the proof of the Proposition for \SimpleSort{}). As for the subtracted term, we have, by definition of $A_{\texttt{sort}(s)}$:
\begin{align*}
& \mathds 1^T A_{\texttt{sort}(s)}^T P_{\texttt{argsort}(s)}\\
= & \mathds 1^T |\texttt{sort}(s) \mathds 1^T - \mathds 1 \texttt{sort}(s)^T| P_{\texttt{argsort}(s)}
\end{align*}
Applying Lemma~\ref{lemma:mult_by_P} to the pointwise absolute value, we can push $P_{\texttt{argsort}(s)}$ into the absolute value:
\begin{align*}
&= \mathds 1^T |\texttt{sort}(s) \mathds 1^TP_{\texttt{argsort}(s)} - \mathds 1 \texttt{sort}(s)^TP_{\texttt{argsort}(s)}|
\end{align*}
Again we can simplify $\texttt{sort}(s)^TP_{\texttt{argsort}(s)} = s^T$ and $\mathds 1^T P_{\texttt{argsort}(s)} = \mathds 1^T$ to get:
\begin{align}\label{eq:almost_done}
&= \mathds 1^T |\texttt{sort}(s) \mathds 1^T - \mathds 1 s^T|
\end{align}
We are almost done. Now just note that we can replace $\texttt{sort}(s)$ in Eq.~\ref{eq:almost_done} by $s$ because multiplication to the left by $\mathds 1^T$ adds up over each column of $|\texttt{sort}(s) \mathds 1^T - \mathds 1 s^T|$ and thus makes the \texttt{sort} irrelevant, hence we get:
\begin{align*}
&= \mathds 1^T |s \mathds 1^T - \mathds 1 s^T|\\
&= \mathds 1^T A_s
\end{align*}
Thus, putting both pieces together into Eq.~\ref{eq:choclo} we arrive at:
\begin{align*}
& = \sigma\Big(\frac{(n + 1 - 2i)s - \mathds 1^T A_{s}}{\tau}\Big) \\
& = \texttt{NeuralSort}_\tau(s)[i, :]
\end{align*}
which proves Proposition~\ref{prop:invariance} for \texttt{NeuralSort}. $\blacksquare$

\section{Proof of Proposition~\ref{dknn}}\label{sec:proof_of_dknn}

First, let us recall the proposition:

\textbf{Proposition}. Let $k=1$ and $\widehat{P}$ be the differentiable kNN operator using $\SimpleSort{}^{|\cdot|}_2$. If we choose the embedding function $\Phi$ to be of norm $1$, then
\[
    \widehat{P}(\hat{y}|\hat{x},X,Y) = 
    {\sum_{i:y_i = \hat{y}} e^{\Phi(\hat{x})\cdot\Phi(x_i)}} \bigg/
    {\sum_{i=1\dots n} e^{\Phi(\hat{x})\cdot\Phi(x_i)}}
\]

\textbf{Proof}.
Since $k=1$, only the first row of the \SimpleSort{} matrix is used in the result. Recall that the elements of the first row are the \texttt{softmax} over $-|s_i - s_{[1]}|$. Given that $s_{[1]} \geq s_i ~\forall i$, we can remove the negative absolute value terms. Because of the invariance of \texttt{softmax} for additive constants, the $s_{[1]}$ term can also be cancelled out.

Furthermore, since the embeddings are normalized, we have that $s_i = -\|\Phi(x_i) - \Phi(\hat{x})\|^2 = 2~\Phi(x_i)\cdot\Phi(\hat{x}) - 2$. When we take the \texttt{softmax} with temperature $2$, we are left with values proportional to $e^{\Phi(x_i)\cdot\Phi(\hat{x})}$. Finally, when the vector is multiplied by $\mathbb{I}_{Y=\hat{y}}$ we obtain the desired identity. $\blacksquare$

\section{Magnitude of Matrix Entries}\label{sec:size}

\begin{figure*}[ht]
\vskip 0.2in
\begin{center}
\centerline{\includegraphics[height=6cm]{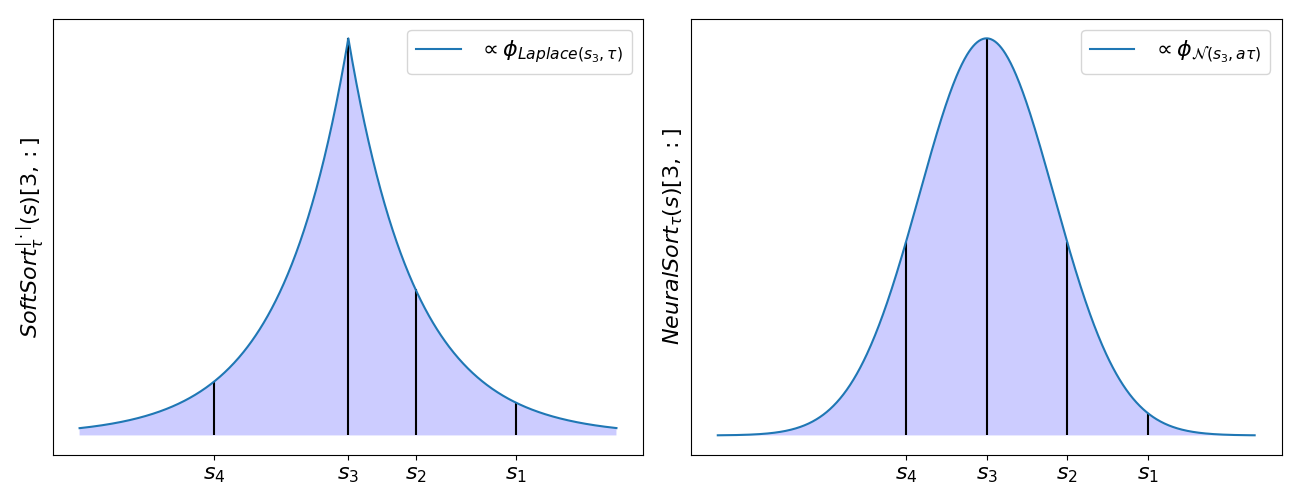}}
\caption{Rows of the $\SimpleSort{}^{|\cdot|}$ operator are proportional to Laplace densities evaluated at the $s_j$, while under the equal-spacing assumption, rows of the \texttt{NeuralSort} operator are proportional to Gaussian densities evaluated at the $s_j$. Similarly, rows of the $\SimpleSort{}^{|\cdot|^2}$ operator are proportional to Gaussian densities evaluated at the $s_j$ (plot not shown).}
\label{fig:laplace_and_gaussian}
\end{center}
\vskip -0.2in
\end{figure*}

The outputs of the \texttt{NeuralSort} and \SimpleSort{} operators are $n\times n$ (unimodal) row-stochastic matrices, i.e. each of their rows add up to one. In section~\ref{sec:mathematical_simplicity} we compared the mathematical complexity of equations~\ref{eq:2018_grover_eq5} and~\ref{eq:simplesort2} defining both operators, but how do these operators differ \textit{numerically}? What can we say about the magnitude of the matrix entries?

For the $\SimpleSort{}^{|\cdot|}$ operator, we show that the values of a given row come from Laplace densities evaluated at the $s_j$. Concretely:

\begin{proposition}\label{prop:simplesort_logits}
For any $s \in \mathbb R^n$, $\tau > 0$ and $1\leq i \leq n$, it holds that $\SimpleSort{}^{|\cdot|}_\tau(s)[i, j] \propto_{j} \phi_{\text{Laplace}(s_\ord{i}, \tau)}(s_j)$. Here $\phi_{\text{Laplace}(\mu, b)}$ is the density of a Laplace distribution with location parameter $\mu \ge 0$ and scale parameter $b > 0$.
\end{proposition}

\textbf{Proof.} This is trivial, since:
\begin{align*}
&\SimpleSort{}^{|\cdot|}_\tau(s)[i, j] = \\
& \underbrace{\frac{1}{\sum_{k = 1}^n \exp\{-|s_\ord{i} - s_k| / \tau\}}}_{c_i} \underbrace{\exp\{-|s_\ord{i} - s_j| / \tau\}}_{\phi_{\text{Laplace}(s_\ord{i}, \tau)}(s_j)}
\end{align*}
where $c_i$ a constant which does not depend on $j$ (specifically, the normalizing constant for row $i$). $\square$

In contrast, for the \texttt{NeuralSort} operator, we show that in the prototypical case when the values of $s$ are equally spaced, the values of a given row of the \texttt{NeuralSort} operator come from \textit{Gaussian} densities evaluated at the $s_j$. This is of course not true in general, but we believe that this case provides a meaningful insight into the \texttt{NeuralSort} operator. Without loss of generality, we will assume that the $s_j$ are sorted in decreasing order (which we can, as argued in section~\ref{sec:mathematical_simplicity}); this conveniently simplifies the indexing. Our claim, concretely, is:

\begin{proposition}\label{prop:neuralsort_logits}
Let $a, b \in \mathbb R$ with $a > 0$, and assume that $s_k = b-ak\ \forall k$. Let also $\tau > 0$ and $i \in \{1,2,\dots,n\}$. Then $\texttt{NeuralSort}_\tau(s)[i, j] \propto_j \phi_{\mathcal{N}(s_i, a\tau)}(s_j)$. Here $\phi_{\mathcal{N}(\mu, \sigma^2)}$ is the density of a Gaussian distribution with mean $\mu \ge 0$ and variance $\sigma^2 > 0$.
\end{proposition}

\textbf{Proof.} The $i,j$-th logit of the \texttt{NeuralSort} operator before division by the temperature $\tau$ is (by Eq.~\ref{eq:2018_grover_eq5}):
\begin{align*}
&(n + 1 - 2i)s_j - \sum_{k = 1}^n |s_k - s_j| \\
&= (n + 1 - 2i)(b-aj) - \sum_{k = 1}^n |b-ak-b+aj| \\
&= (n + 1 - 2i)(b-aj) - a\sum_{k = 1}^n |j - k| \\
&= (n + 1 - 2i)(b-aj) - a\frac{j(j - 1)}{2} \\
&\hspace{10em}  - a \frac{(n - j)(n - j + 1)}{2} \\
&= -a(i - j)^2 + a(i^2 - \frac{n^2}{2} - \frac{n}{2}) -b(2i - n - 1) \\
&= -\frac{(s_i - s_j)^2}{a} + \underbrace{a(i^2 - \frac{n^2}{2} - \frac{n}{2}) -b(2i - n - 1)}_{\Delta_i}
\end{align*}
where $\Delta_i$ is a constant that does not depend on $j$. Thus, after dividing by $\tau$ and taking softmax on the $i$-th row, $\Delta_i / \tau$ vanishes and we are left with:
\begin{align*}
&\texttt{NeuralSort}_\tau[i, j] = \\
& \underbrace{\frac{1}{\sum_{k = 1}^n \exp\{-(s_{i} - s_k)^2 / (a\tau)\}}}_{c_i} \underbrace{\exp\{-(s_{i} - s_j)^2 / (a\tau)\}}_{\phi_{\mathcal N(s_i, a\tau)}(s_j)}
\end{align*}
where $c_i$ a constant which does not depend on $j$ (specifically, the normalizing constant for row $i$). $\square$

Gaussian densities can be obtained for \SimpleSort{} too by choosing $d = |\cdot|^2$. Indeed:

\begin{proposition}\label{prop:simplesort_logits_p2}
For any $s \in \mathbb R^n$, $\tau > 0$ and $1\leq i \leq n$, it holds that $\SimpleSort{}^{|\cdot|^2}_\tau(s)[i, j] \propto_{j} \phi_{\mathcal N(s_\ord{i}, \tau)}(s_j)$.
\end{proposition}

\textbf{Proof.} This is trivial, since:
\begin{align*}
&\SimpleSort{}^{|\cdot|^2}_\tau(s)[i, j] = \\
& \underbrace{\frac{1}{\sum_{k = 1}^n \exp\{-(s_\ord{i} - s_k)^2 / \tau\}}}_{c_i} \underbrace{\exp\{-(s_\ord{i} - s_j)^2 / \tau\}}_{\phi_{\mathcal N(s_\ord{i}, \tau)}(s_j)}
\end{align*}
where $c_i$ a constant which does not depend on $j$ (specifically, the normalizing constant for row $i$). $\square$

Figure~\ref{fig:laplace_and_gaussian} illustrates propositions~\ref{prop:simplesort_logits} and~\ref{prop:neuralsort_logits}. As far as we can tell, the Laplace-like and Gaussian-like nature of each operator is neither an advantage nor a disadvantage; as we show in the experimental section, both methods perform comparably on the benchmarks. Only on the speed comparison task does it seem like \texttt{NeuralSort} and $\SimpleSort{}^{|\cdot|^2}$ outperform $\SimpleSort{}^{|\cdot|}$.

Finally, we would like to remark that $\SimpleSort{}^{|\cdot|^2}$ does not recover the \texttt{NeuralSort} operator, not only because Proposition~\ref{prop:neuralsort_logits} only holds when the $s_i$ are equally spaced, but also because even when they \textit{are} equally spaced, the Gaussian in Proposition~\ref{prop:neuralsort_logits} has variance $a\tau$ whereas the Gaussian in Proposition~\ref{prop:simplesort_logits_p2} has variance $\tau$. Concretely: we can only make the claim that $\SimpleSort{}^{|\cdot|^2}_{a\tau}(s) = \texttt{NeuralSort}_\tau(s)$ when $s_i$ are equally spaced at distance $a$. As soon as the spacing between the $s_i$ changes, we need to \textit{change the temperature} of the $\SimpleSort{}^{|\cdot|^2}$ operator to match the \texttt{NeuralSort} operator again. Also, the fact that the $\SimpleSort{}^{|\cdot|^2}$ and \texttt{NeuralSort} operators agree in this prototypical case for some choice of $\tau$ does not mean that their \textit{gradients} agree. An interesting and under-explored avenue for future work might involve trying to understand how the \textit{gradients} of the different continuous relaxations of the \texttt{argsort} operator proposed thus far compare, and whether some gradients are preferred over others. So far we only have empirical insights in terms of learning curves.

\section{Sorting Task - Proportion of Individual Permutation Elements Correctly Identified}\label{app:run_sort_aux}

Table~\ref{tab:run_sort_supp} shows the results for the second metric (the proportion of individual permutation elements correctly identified). Again, we report the mean and standard deviation over 10 runs. Note that this is a less stringent metric than the one reported in the main text. The results are analogous to those for the first metric, with \SimpleSort{} and \texttt{NeuralSort} performing identically for all $n$, and outperforming the method of \cite{2019_cuturi} for $n = 9, 15$.

\begin{table*}[t]
\caption{Results for the sorting task averaged over 10 runs. We report the mean and standard deviation for the \textit{proportion of individual permutation elements correctly identified}.}
\label{tab:run_sort_supp}
\vskip 0.15in
\begin{center}
\begin{small}
\begin{sc}
\resizebox{2.0\columnwidth}{!}{
\begin{tabular}{lccccc}
\toprule
Algorithm & $n = 3$ & $n = 5$ & $n = 7$ & $n = 9$ & $n = 15$ \\
\midrule
Deterministic NeuralSort & $0.946\ \pm\ 0.004$ & $0.911\ \pm\ 0.005$ & $\textbf{0.882}\ \pm\ \textbf{0.006}$ & $\textbf{0.862}\ \pm\ \textbf{0.006}$ & $\textbf{0.802}\ \pm\ \textbf{0.009}$ \\
Stochastic NeuralSort & $0.944\ \pm\ 0.004$ & $0.912\ \pm\ 0.004$ & $\textbf{0.883}\ \pm\ \textbf{0.005}$ & $\textbf{0.860}\ \pm\ \textbf{0.006}$ & $\textbf{0.803}\ \pm\ \textbf{0.009}$ \\
\midrule
Deterministic SoftSort & $0.944\ \pm\ 0.004$ & $0.910\ \pm\ 0.005$ & $\textbf{0.883}\ \pm\ \textbf{0.007}$ & $\textbf{0.861}\ \pm\ \textbf{0.006}$ & $\textbf{0.805}\ \pm\ \textbf{0.007}$ \\
Stochastic SoftSort & $0.944\ \pm\ 0.003$ & $0.910\ \pm\ 0.002$ & $\textbf{0.884}\ \pm\ \textbf{0.006}$ & $\textbf{0.862}\ \pm\ \textbf{0.008}$ & $\textbf{0.802}\ \pm\ \textbf{0.007}$ \\
\midrule
\citep[reported]{2019_cuturi} & $\textbf{0.950}$ & $\textbf{0.917}$ & $\textbf{0.882}$ & $0.847$ & $0.742$ \\
\bottomrule
\end{tabular}
}
\end{sc}
\end{small}
\end{center}
\vskip -0.1in
\end{table*}

\section{PyTorch Implementation}

In Figure~\ref{code:simplesort_pytorch} we provide our PyTorch implementation for the $\SimpleSort{}^{|\cdot|}$ operator. Figure~\ref{code:neuralsort_pytorch} shows the PyTorch implementation of the \texttt{NeuralSort} operator \cite{2018_grover} for comparison, which is more complex.

\begin{figure*}[ht]
\vskip 0.2in
\begin{center}
\begin{verbatim}
def neural_sort(s, tau):
    n = s.size()[1]
    one = torch.ones((n, 1), dtype = torch.float32)
    A_s = torch.abs(s - s.permute(0, 2, 1))
    B = torch.matmul(A_s, torch.matmul(one, torch.transpose(one, 0, 1)))
    scaling = (n + 1 - 2 * (torch.arange(n) + 1)).type(torch.float32)
    C = torch.matmul(s, scaling.unsqueeze(0))
    P_max = (C-B).permute(0, 2, 1)
    sm = torch.nn.Softmax(-1)
    P_hat = sm(P_max / tau)
    return P_hat
\end{verbatim}
\caption{Implementation of \texttt{NeuralSort} in PyTorch as given in \cite{2018_grover}}
\label{code:neuralsort_pytorch}
\end{center}
\vskip -0.2in
\end{figure*}

\begin{figure*}[ht]
\vskip 0.2in
\begin{center}
\begin{verbatim}
def soft_sort(s, tau):
    s_sorted = s.sort(descending=True, dim=1)[0]
    pairwise_distances = (s.transpose(1, 2) - s_sorted).abs().neg() / tau
    P_hat = pairwise_distances.softmax(-1)
    return P_hat
\end{verbatim}
\caption{Implementation of \SimpleSort{} in PyTorch as proposed by us (with $d = |\cdot|$).}
\label{code:simplesort_pytorch}
\end{center}
\vskip -0.2in
\end{figure*}